\documentclass[10pt,twocolumn,letterpaper, pagebackref,breaklinks,colorlinks]{article}

\usepackage[pagenumbers]{cvpr}
\usepackage{cvpr}
\usepackage{url}
\usepackage{graphicx}
\usepackage{amsmath}
\usepackage{amssymb}
\usepackage{booktabs}
\usepackage{xcolor}
\usepackage{multirow}
\usepackage{wrapfig}
\usepackage{float}
\usepackage{bbding}
\usepackage[linesnumbered,ruled,vlined]{algorithm2e}
\usepackage{algorithmic,algorithm2e,float}
\usepackage{bm}
\usepackage{color, colortbl}
\usepackage{float}

\usepackage{color}
\usepackage[pagebackref=true,breaklinks=true,letterpaper=true,colorlinks,citecolor=citecolor,bookmarks=false]{hyperref}
\definecolor{citecolor}{RGB}{30,130,255}
\definecolor{palegreen}{RGB}{210,240,200}

\usepackage[capitalize]{cleveref}
\usepackage[misc]{ifsym}

\SetKwInput{KwInput}{Input}
\SetKwInput{KwOutput}{Output} 
\hypersetup{colorlinks=true,linkcolor=blue, linktocpage}
\definecolor{Green}{rgb}{0.85882353, 0.90980392, 0.84705882}

\crefname{section}{Sec.}{Secs.}
\Crefname{section}{Section}{Sections}
\Crefname{table}{Table}{Tables}
\crefname{table}{Tab.}{Tabs.}

\newcommand{\redscore}[2]{$\text{#1}^{\text{\footnotesize{\color{red}{#2}}}}$}
\newcommand{\greenscore}[2]{$\text{#1}^{\text{\footnotesize{\color{green}{#2}}}}$}
\newcommand{\corrAuthor}{$^{\textrm{\Letter}}$}

\def\cvprPaperID{7803}
\def\confName{CVPR}
\def\confYear{2023}

\begin{document}

\title{PromptCAL: Contrastive Affinity Learning via Auxiliary Prompts for Generalized Novel Category Discovery}

\author{Sheng Zhang$^{\dagger}$\corrAuthor \quad Salman Khan$^{\dagger\diamond}$ \quad Zhiqiang Shen$^{\dagger \star}$ \quad Muzammal Naseer$^{\dagger}$ \\
 Guangyi Chen$^{\dagger\ddag}$ \quad Fahad Khan$^{\dagger\ast}$\\
$^{\dagger}$Mohamed bin Zayed University of AI \quad $^{\star}$Hong Kong University of Science and Technology  \\
$^{\diamond}$Australian National University \quad $^{\ast}$Link\"{o}ping University \quad $^\ddag$Carnegie Mellon University \\
{\tt \small \{firstname.lastname\}@mbzuai.ac.ae}
}
\maketitle

\renewcommand*\ttdefault{cmvtt}
\begin{abstract}
Although existing semi-supervised learning models achieve remarkable success in learning with unannotated in-distribution data, they mostly fail to learn on unlabeled data sampled from novel semantic classes due to their closed-set assumption. 
In this work, we target a pragmatic but under-explored {Generalized Novel Category Discovery} (GNCD) setting. 
The GNCD setting aims to categorize unlabeled training data coming from known and novel classes by leveraging the information of partially labeled known classes.
We propose a two-stage Contrastive Affinity Learning method with auxiliary visual Prompts, dubbed \emph{\mbox{PromptCAL}}, to address this challenging problem. 
Our approach discovers reliable pairwise sample affinities to learn better semantic clustering of both known and novel classes for the class token and visual prompts. 
First, we propose a discriminative prompt regularization loss to reinforce semantic discriminativeness of prompt-adapted pre-trained vision transformer for refined affinity relationships.
Besides, we propose contrastive affinity learning to calibrate semantic representations based on our iterative semi-supervised affinity graph generation method for semantically-enhanced supervision.
Extensive experimental evaluation demonstrates that our PromptCAL method is more effective in discovering novel classes even with limited annotations and surpasses the current state-of-the-art on generic and fine-grained benchmarks (\eg, with nearly $11\%$ gain on CUB-200, and $9\%$ on ImageNet-100) on overall accuracy. 
Our code is available at \url{https://github.com/sheng-eatamath/PromptCAL}.
\end{abstract}
\vspace{-10pt}

\section{Introduction}
\label{sec:intro}
The deep neural networks have demonstrated favorable performance in the \textit{Semi-Supervised Learning} (SSL) setting~\cite{ssl-survey, s4l, noisy-student, cnn-recognition, comatch}.
Some recent works can even achieve comparable performance to their fully-supervised counterparts using few annotations for image recognition~\cite{mt, noisy-student, semi-vit}.
However, these approaches heavily rely on the \emph{closed-world} assumption that unlabeled data share the same underlying class label space as the labeled data~\cite{ood-survey, os-survey}. 
In many realistic scenarios, this assumption does not hold true because of the inherent dynamism of real-world tasks where novel classes can emerge in addition to known classes. \par 

\begin{figure}[t]
    \centering
    \includegraphics[scale=0.35]{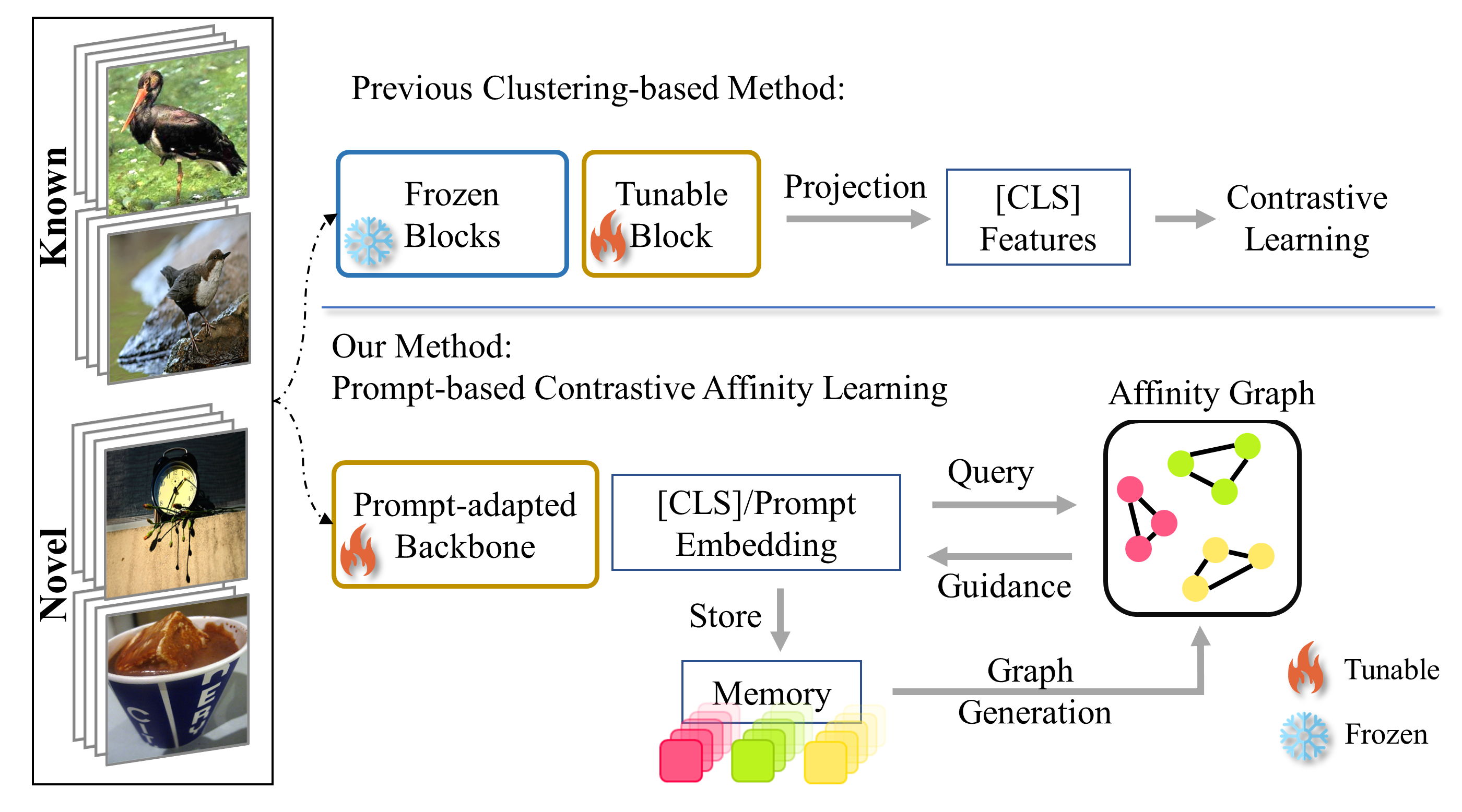}
    \vspace{-10pt}
    \caption{\textbf{PromptCAL Overview.} In contrast to previous method based on semi-supervised contrastive learning, PromptCAL constructs affinity graph on-the-fly to guide representation learning of the class token and prompts. Meanwhile, our prompt-adapted backbone can be tuned to enhance semantic discriminativeness.
    PromptCAL can discover reliable affinities from a memory bank, especially for novel classes.
    Therefore, our PromptCAL is better task-aligned and discriminative to novel semantic information.
    }
    \vspace{-20pt}
    \label{fig:advertise}
\end{figure}

In contrast to SSL, the \textit{Novel Category Discovery}~(NCD) problem was introduced by \cite{dtc} to relax the closed-world assumption of SSL, which assumes the unlabeled data contain novel classes. 
Recently, the nascent \textit{Generalized Novel Category Discovery}~(GNCD) problem, first proposed in \cite{gcd, orca}, extends NCD and assumes the unlabeled data can contain both known and novel classes, which is more pragmatic and challenging.
To be more specific, GNCD intends to categorize images sampled from predefined categories in the training set comprising labeled-knowns, unlabeled-knowns, and unlabeled-novels. \par 
Our work focuses on GNCD problem. The key challenge of GNCD is to discriminate among novel classes when only the ground truths of known classes are accessible in the training set. 
Recent studies show that self-supervised pre-trained representations are conducive to discovering novel semantics
\cite{deep-cluster, gcd, orca, ncl, rs}.
A typical work on GNCD \cite{gcd} takes advantage of the large-scale pre-trained visual transformer (ViT) \cite{vit}, and learns robust clusters for known and novel classes through semi-supervised contrastive learning on downstream datasets. 
However, we discover that the remarkable potential of pre-trained ViT is actually suppressed by this practice, due to the \emph{class collision}~\cite{wscl} issue induced by abundant false negatives in contrastive loss, \ie, considering different unlabeled images from the same or similar semantic class as false negatives. As supported by empirical studies, abundant false negatives in contrastive training can deteriorate the compactness and purity of semantic clustering~\cite{fnc, slic, wscl, deep-cluster}. Based on empirical investigation, we show that this issue is particularly severe in category discovery.
Furthermore, although the existing commonly adopted practice \cite{gcd, orca} of freezing most parts of the pre-trained backbone can alleviate overfitting on known classes, it constrains the flexibility and adaptability of backbones \cite{vpt}. Lack of adaptability inhibits models from learning discriminative semantic information on downstream datasets. \par

To address above limitations and learn better semantically discriminative representations, we propose \textbf{Prompt}-based \textbf{C}ontrastive \textbf{A}ffinity \textbf{L}earning (\textbf{\mbox{PromptCAL}}) framework to tackle GNCD problem.
To be specific, our approach aims to discover semantic clusters in unlabeled data by simultaneous semantic prompt learning based on our Discriminative Prompt Regularization (DPR) loss and representation calibration based on our Contrastive Affinity Learning (CAL) process.
\emph{Firstly}, CAL discovers abundant reliable pseudo positives for DPR loss and contrastive loss based on generated affinity graphs. These semantic-aware pseudo labels further enhance the semantic discriminativeness of DPR supervision.
\emph{Secondly}, DPR regularizes semantic representations of ensembled prompts, which facilitates the discovery of more accurate pseudo labels at the next-step of CAL.
Therefore, as model and prompt representations are iteratively enhanced, we can obtain higher quality pseudo positives for further self-training as well as acquire better semantic clustering. \par

Our PromptCAL achieves State-Of-The-Art (SOTA) performance in extensive experimental evaluation on six benchmarks. Specifically, PromptCAL remarkably surpasses previous SOTA by more than $10\%$ clustering accuracy on the fine-grained CUB-200 and StandfordCars datasets; it also significantly outperforms previous SoTAs by nearly $4\%$ on ImageNet-100 and $7\%$ on CIFAR-100. 
Interestingly, we identify that both DPR supervised prompts and unsupervised prompts of PromptCAL can learn semantic discriminativeness, which advances the flexibility of the pre-trained backbone. 
Furthermore, PromptCAL still achieves the best performance in challenging low-labeling and few-class setups. \par

\noindent \textbf{ Our contributions} include: \textbf{(1)} We propose a two-stage framework for the generalized novel category discovery problem, in which semantic prompt tuning and contrastive affinity learning mutually reinforce and benefit each other during the learning process.
\textbf{(2)} We propose two synergistic learning objectives, contrastive affinity loss and discriminative prompt regularization loss, based on our semi-supervised adapted affinity graphs to enhance semantic discriminativeness.
\textbf{(3)} We comprehensively evaluate our method on three generic (\ie, CIFAR-10, CIFAR-100, and ImageNet-100) and three fine-grained benchmarks (\ie, CUB-200, Aircraft, and StandfordCars), achieving state-of-the-art performance, thereby showing its effectiveness.
\textbf{(4)} We further showcase generalization ability of PromptCAL and its effectiveness in more challenging low-labeling and few-class setups. \par

\section{Related Work}
\label{sec:rw}
\noindent \textbf{Category discovery.}
\textit{Novel Category Discovery} (NCD), first formulated by DTC~\cite{dtc}, aims to categorize the unlabeled novel classes by transferring the knowledge from labeled known classes~\cite{dtc, rs,rs++, openmix, dec, ncl, uno}. 
The challenging NCD differs from SSL~\cite{ssl-survey} in that the unlabeled data are sampled from distinct underlying semantic distribution.
DTC~\cite{dtc} proposes to jointly warm up network weights and cluster prototypes based on DEC~\cite{dec} method on unlabeled data, and then fit an annealing sharpened distribution.
RankStats~\cite{rs} and RS+~\cite{rs++} propose to utilize ranking statistics to generate pseudo positives among unlabeled novel classes.
OpenMix~\cite{openmix} transfers semantic knowledge by MixUp augmentation~\cite{mixup} between known and novel classes as well as between reliable novel anchors and other novel examples.
NCL~\cite{ncl} proposes a neighborhood contrastive loss and a hard-negative generation process by mixing~\cite{mixup} novel and known classes.
UNO~\cite{uno} first formulates the NCD problem into classification based on dynamic class assignments by Sinkhorn-Knopp algorithm~\cite{sk}. 
WTA~\cite{wta} addresses multi-modal novel category discovery by inter- and intra-modal contrastive learning with permutation-ensembled ranking statistics as the pseudo-labeling method. \par 

\textit{Generalized Novel Category Discovery} (GNCD) problem, first proposed in~\cite{gcd}, further extends NCD under a more realistic assumption that unlabeled data can be both sampled from novel classes and known classes. Specifically, the model learns to categorize unlabeled training data containing known and novel classes based on the knowledge of labeled known classes.
Besides, a concurrent work, ORCA~\cite{orca} proposes an uncertainty adaptive margin loss to reduce the intra-class variances between known and novel classes.
GCD~\cite{gcd} addresses this challenging problem via proposed semi-supervised contrastive learning on large-scale pre-trained visual transformer (ViT) followed by constraint KMeans~\cite{kmeanspp, ckmeans}.
However, GCD still has limitations: first, the frozen backbone lacks the adaptability to downstream tasks; besides, abundant false negatives will degenerate the semantic representation~\cite{fnc, ifnd, wscl, slic}.
To fully unleash the potential of pre-trained ViT, we address these two critical issues via our proposed prompt-based contrastive affinity learning. \par

\noindent \textbf{Positive Mining in Neighborhoods.}
Some recent works in self-supervised learning discovered that mining positives to antagonize the side effect of abundant false negatives in the sample-wise contrastive loss is essential to the downstream performance~\cite{slic, little-help, fnc, wscl, finch}.
FNC~\cite{fnc} comprehensively analyzes the adverse effect of false-negatives on contrastive learning SoTAs and performs positive mining based on ensembled similarities on local patch pairs.
LA~\cite{local-agg} proposes to learn better representation through soft clusters in neighborhoods at different scales.
NNCLR~\cite{little-help}, NCL~\cite{ncl}, and WCL~\cite{wscl} conduct positive mining based on K-Nearest Neighbors (KNN) as pseudo positives to improve contrastive learning.
We find one work in SSL \cite{semi-prop} also leverages a graph diffusion algorithm to propagate pseudo labels.
But there exist major differences between their work and ours: first, features in our context are prone to open-set noises~\cite{os-survey} and thus more challenging than SSL; second, we conduct an efficient online diffusion per iteration via a graph subsampling strategy, while they conduct diffusion per epoch on the entire dataset; third, we compute affinity propagation on consensus affinity graph with prior knowledge, while they conduct propagation on naive KNN graph.
Our framework incorporates and generalizes consensus KNN~\cite{consensus-knn}, which was originally built upon non-learnable SIFT~\cite{sift} features of synthetic datasets, while our method exploits deep features and can guide end-to-end training, which suits the GNCD context. \par

\noindent \textbf{Visual prompt learning.}
Prompt learning originates from the field of Natural Language Processing (NLP)~\cite{nlp-prompt-survey}.
Visual prompt learning (VPT)~\cite{vpt} tunes embedded visual prompts with a frozen pre-trained ViT backbone supervised by downstream objectives, which achieves better transfer.
However, based on our experimental analysis, VPT~\cite{vpt} does not exhibit significant benefits especially on fine-grained datasets.
Our objective acts as prompt regularization or a weaker semantic supervision signal, which is distinct from the learning goals of prompt ensembling~\cite{prompt-ensemble, prompt-ensemble-1} and prompt composition~\cite{ptr} in NLP~\cite{nlp-prompt-survey}. \par

\begin{figure*}
    \centering
    \includegraphics[scale=0.4]{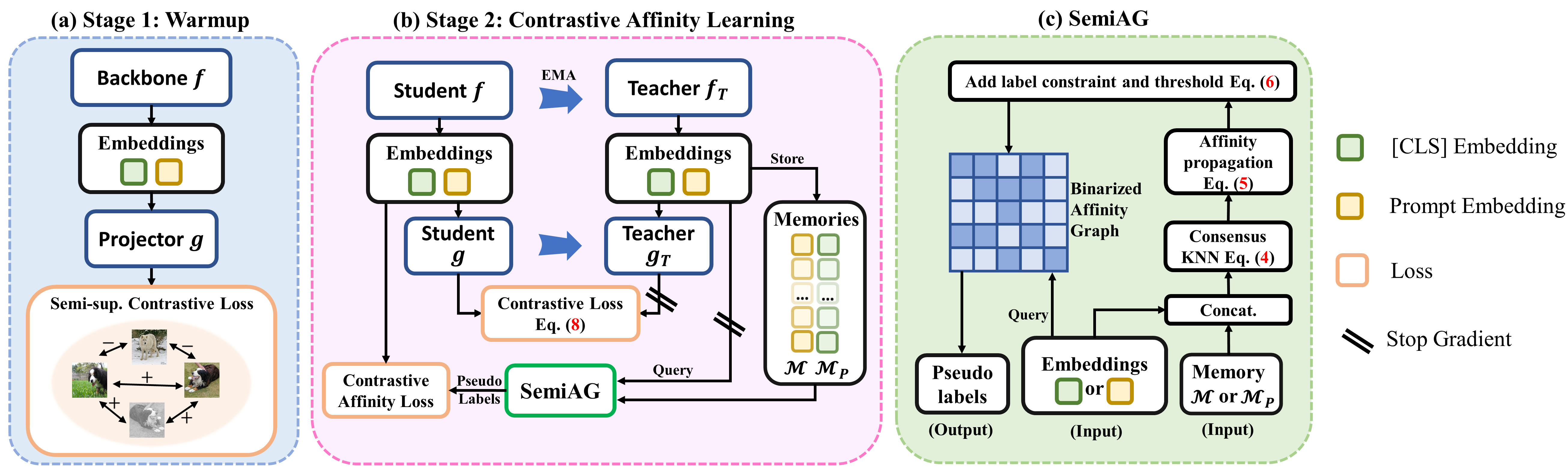}
    \vspace{-18pt}
    \caption{\textbf{Overview of our PromptCAL framework.} 
    Our prompt-adapted backbone outputs a class embedding and ensembled prompt embedding.
    (a) In warm-up training, we conduct semi-supervised contrastive clustering (Semi-sup. Contrastive Loss) on the projected features of the class token and ensembled prompt, respectively.
    (b) In contrastive affinity learning stage, at each iteration, we forward the student and EMA (exponentially moving averaged) teacher with different augmented views of images. Output teacher embeddings are enqueued into their corresponding token-specific memory.
    We iteratively compute semi-supervised contrastive loss on the current batch and our contrastive affinity loss for student embeddings and memory embeddings with pseudo-labels from the dynamically generated affinity graph by SemiAG.
    (c) We generate affinity graphs for the class embedding and prompt embedding respectively via affinity propagation with label constraints on their corresponding consensus KNN graphs.
    }
    \vspace{-10pt}
    \label{fig:main}
\end{figure*}

\section{Method}
\label{sec:method}
The challenging aspect of GNCD in comparison to SSL is clustering novel semantics under both semantic shifts and missing annotations~\cite{ood-survey, ssl-survey}.
However, existing methods~\cite{gcd, rs, rs++, ncl} cannot reliably discover and employ semantic affinities on pre-trained representations.
Meanwhile, recent SoTAs~\cite{gcd, orca} lack suitable strategies to adapt the pre-trained backbone to learn discriminative semantic information without overfitting on known classes. \par

To this end, we propose \mbox{PromptCAL}, which consists of two synergistic learning objectives: discriminative prompt regularization (DPR) and contrastive affinity learning (CAL).
The whole framework is displayed in Fig.~\ref{fig:main}.
Specifically, in the first stage, we learn warm-up representation (in Sec.~\ref{method:warm-up}) for further tuning. Our DPR loss which is applied to both stages for prompt regularization is also explained.
In the second stage, we discover reliable pseudo positives on generated affinity embedding graphs based on semi-supervised affinity generation (SemiAG) mechanism (in Sec.~\ref{method:semi-ap}). 
Next, we propose our contrastive affinity loss (in Sec.~\ref{method:cal}) on pseudo labels generated by online SemiAG with the support of embedding memories.
Lastly, we also present \mbox{PromptCAL} training algorithm in Appendix~E. \par

\subsection{Preliminaries}
\label{method:formulation}
Before introducing our method, we formulate the GNCD problem and present some preliminaries. \par 

\noindent \textbf{Problem Definition.} 
Our GNCD setting follows \cite{gcd}. Specifically, we assume that the training dataset $\mathcal{D}=\mathcal{D}_l \bigcup \mathcal{D}_u$ comprises two subsets: a labeled set $\mathcal{D}_l=\{\pmb{x}_i, y_i\}_{i=1}^{N_1} \subset \mathcal{X}_l \times \mathcal{Y}_l$ with its label space $\mathcal{Y}_l=\mathcal{C}_{kwn}$, and an unlabeled set $\mathcal{D}_u=\{\pmb{x}_i\}_{i=1}^{N_2} \subset \mathcal{X}_u$ with its underlying label space $\mathcal{Y}_u=\mathcal{C}=\mathcal{C}_{kwn} \bigcup \mathcal{C}_{new}$. 
Here, $\mathcal{C}$, $\mathcal{C}_{kwn}$, and $\mathcal{C}_{new}$ denote the label set for {\tt All}, {\tt Known}, and {\tt New} classes, respectively. Following~\cite{gcd}, we assume $|\mathcal{C}|$ is known. \par

\noindent \textbf{Architecture.} We take a self-supervised pre-trained ViT as our backbone~\cite{vit}.
We denote our visual prompt-adapted ViT backbone~\cite{vpt} as $f(\cdot | \theta, \theta_{\text{P}})$ parameterized by prompts $\theta_{\text{P}}$ and last block weights $\theta$. 
In each mini-batch $\mathcal{B}$, there are two augmented views for each sample.
Given a sample vector $\mathbf{x} \in \mathcal{B}$, we can extract its embedding $\mathbf{h}=f(\mathbf{x} | \theta, \theta_{\text{P}}) \in \mathcal{H}$ and project $\mathbf{h}$ into feature vector $\mathbf{z}=g(\mathbf{h} | \theta_{\text{H}}) \in \mathcal{Z}$ through a projection head $g(\cdot | \theta_{\text{H}})$ with parameters $\theta_\text{H}$. 
Here, $\mathcal{H}, \mathcal{Z}$ denote embedding and feature spaces. \par

\noindent \textbf{Contrastive Loss.} To simplify notations of \mbox{PromptCAL}, we extend the definition of the standard supervised contrastive loss~\cite{scl} as follows. 
Given a $l_2$-normalized query vector $\mathbf{t}_q$ and a set of $l_2$-noramlized key vectors $\mathbf{T}_k$ (which can be from the embedding or feature space), we define:
\begin{equation}
\begin{aligned}
 & L_{\text{con}}(\mathbf{t}_q, \mathbf{T}_k; \tau, \mathcal{P}, \mathcal{A}) \\
 & = - \frac{1}{|\mathcal{P}(\mathbf{t}_q)|} \sum_{\mathbf{t}_k^+ \in \mathcal{P}(\mathbf{t}_q)}{\frac{\exp(\frac{\mathbf{t}_q \cdot \mathbf{t}_k^+}{\tau})}{ \sum_{\mathbf{t_a} \in \mathcal{A}( \mathbf{t}_q ) }\exp(\frac{\mathbf{t}_q \cdot \mathbf{t_a}}{\tau})}} \label{eq:gcl}
\end{aligned}
\end{equation}
where $\tau$ is the temperature parameter of the contrastive loss, and $\cdot$ denotes the cosine similarity operation. 
Here, $\mathcal{P}(\mathbf{t}_q)$ and $\mathcal{A}(\mathbf{t}_q)$ represent the positive set and anchor set of the query $\mathbf{t}_q$, respectively, which are subsets of $\mathbf{T}_k$. \par

\subsection{Warm-up Phase with Discriminative Prompt Regularization}
\label{method:warm-up}
\noindent \textbf{Discriminative Prompt Regularization.} Although computation overheads are largely reduced by only tuning the last block~\cite{gcd}, it restricts the backbone from better learning semantic representations and adapting to diverse downstream datasets.
Counterintuitively, we discover that naively adapting backbone with visual prompts~\cite{vpt} overfits small datasets (refer to ablations on CUB-200~\cite{dataset-cub} in Sec.~\ref{exp:ablation}). \par

Motivated by~\cite{dense-sup, filter-bank}, we propose a discriminative prompt regularization loss to regularize and force prompts to learn semantically discriminative features with a task-related auxiliary loss.
We investigate the superiority of DPR supervision on our prompt-adapted backbone in ablation study (Sec.~4.5) and Appendix D. \par

We assign input prompts at the last ViT block as {\tt [P]} tokens (short for prompt), the output of which are ensembled and supervised by a task-related clustering loss in both training stages. 
All the remaining prompts are unsupervisedly learned, which provides the backbone with extra flexibility. 
Concretely, we average the $l_2$-normalized output embeddings of all {\tt [P]} tokens into an ensembled embedding $\mathbf{h}_{\text{P}}$ (the same shape as the class embedding), and forward it to the projection head and obtain $\mathbf{z}_{\text{P}}$. Finally, we define the DPR task-related loss function on $\mathbf{h}_{\text{P}}$/$\mathbf{z}_{\text{P}}$ as the same form of the loss defined on $\mathbf{h}$/$\mathbf{z}$ but with a weaker weight $\gamma$. \par

\noindent \textbf{Warm-up Training.} Since randomly initialized prompts are not ready for contrastive affinity learning, we include warm-up training to prepare the class token and prompts with dataset-specific representation.
The overall training objective in this stage is formulated as:
\begin{equation}
    L_1(\mathbf{x}) = L_{\text{semi}}^{\text{CLS}}(\mathbf{z}) + \gamma L_{\text{semi}}^{\text{P}}(\mathbf{z}_{\text{P}}) \label{eq:warm-up}
\end{equation}
where $L_{\text{semi}}^{\text{CLS}}$ and $L_{\text{semi}}^{\text{P}}$ represent the semi-supervised contrastive loss (SemiCL) on {\tt [CLS]} and its DPR counterpart on {\tt [P]}, respectively. Here, $\gamma$ is DPR loss weight.
Further, based on extended contrastive loss (Eq.~\ref{eq:gcl}), the SemiCL on {\tt [CLS]} feature $\mathbf{z} \in \mathbf{Z}_{\mathcal{B}}$ is written as:
\begin{equation}
\begin{aligned}
    L_{\text{semi}}^{\text{CLS}}(\mathbf{z}) &= (1- \alpha) L_{\text{con}}\Big(\mathbf{z}, \mathbf{Z}_{\mathcal{B}}; \tau, \mathcal{P}_{\text{self}}, \mathcal{A}_{\text{self}} \Big) \\
    &+ \alpha L_{\text{con}}\Big(\mathbf{z}, \mathbf{Z}_{\mathcal{B}_l}; \tau_a, \mathcal{P}_{\text{sup}}, \mathcal{A}_{\text{sup}}\Big) \mathbb{I}\Big(\mathbf{z} \in \mathbf{Z}_{\mathcal{B}_l} \Big)
    \label{eq:semi-cl}
\end{aligned}
\end{equation}
where $\tau, \tau_a$ are temperature parameters, and $\mathbb{I}$ is an indicator function. The first and second terms denote self-supervised and supervised contrastive loss on projected features of an entire batch $\mathbf{Z_{\mathcal{B}}}$ and only labeled samples $\mathbf{Z_{\mathcal{B}_l}}$, respectively.
Following~\cite{moco, scl}, we define $\mathcal{P}_{\text{self}}(\mathbf{z})$ as the augmented counterpart of $\mathbf{z}$ in $\mathbf{Z_{\mathcal{B}}}$, and define $\mathcal{P}_{\text{sup}}(\mathbf{z})$ as all other features in $\mathbf{Z_{\mathcal{B}_l}}$ that shares the same class label with $\mathbf{z}$. Besides, we have $\mathcal{A}_{\text{sup}}(\mathbf{z})=\mathbf{Z_{\mathcal{B}_l}}-\{\mathbf{z}\}$ and $\mathcal{A}_{\text{self}}(\mathbf{z})=\mathbf{Z_{\mathcal{B}}}-\{\mathbf{z}\}$.
Similar to Eq.~\ref{eq:semi-cl}, we can define the DPR loss function $L_{\text{semi}}^{\text{P}}$ on ensembled prompt feature $\mathbf{z}_{\text{P}}$ in the overall loss (Eq.~\ref{eq:warm-up}). \par

\begin{figure*}[htp]
    \centering
    \includegraphics[scale=0.6]{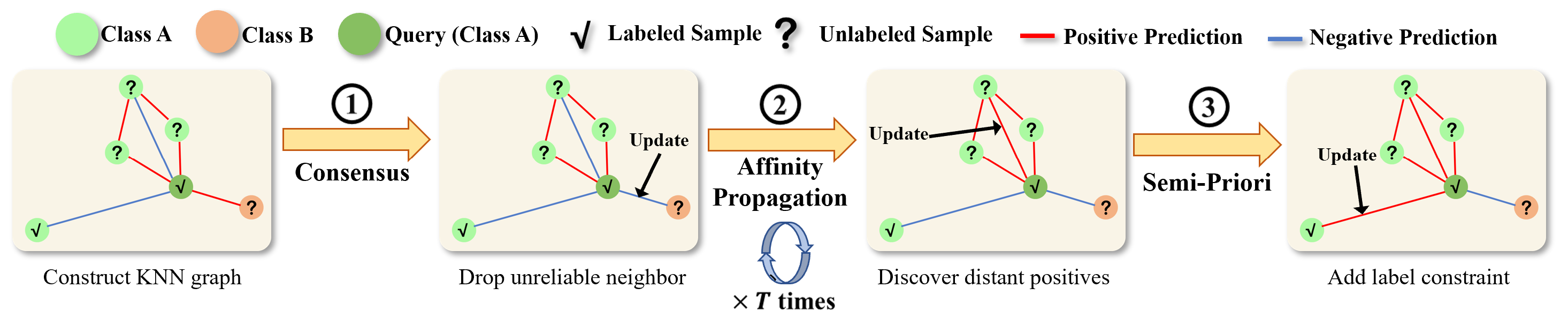}
    \vspace{-10pt}
    \caption{\textbf{An intuitive toy example for SemiAG}, which sequentially requires three operations. 
    The relative pairwise distances are proportional to cosine distances in the embedding space.
    Each of the four graphs denotes results obtained at each step after binarization with thresholds.
    Each operation can either remove false positives or retrieve ground-truth positives for the query embedding (dark green). 
    Firstly, only reliable neighbors are retrieved as positives based on consensus information; secondly, more positives are retrieved by affinity propagation on the entire graph; and thirdly, pairwise constraints in label information of labeled data (SemiPriori) are incorporated for affinity calibration.
    }
    \vspace{-10pt}
    \label{fig:semiag}
\end{figure*}

\subsection{Semi-supervised Affinity Generation}
\label{method:semi-ap}
Once the warmed-up semantic representation for the class token and prompts are obtained, abundant positive samples can be discovered by reliable pseudo-labeling methods for enhanced clustering and supervision signals at next iteration.
However, pseudo-labeling techniques in recent works (\eg, naive nearest neighbors, pair-wise predictions as positives~\cite{ncl, orca, uno, rs, slic}) are not robust enough to semantic shifts~\cite{semi-eval}. 
To address this issue, we propose a semi-supervised affinity generation method under the assumption that consensus local neighbors share the same semantics.
Specifically, we \emph{first} construct an consensus affinity graph in $\mathcal{H}$ based on neighborhood statistics~\cite{consensus-knn}. 
\emph{Then}, we conduct affinity propagation on the entire graph to calibrate affinities.
\emph{Lastly}, we incorporate the semi-supervised priori from $\mathcal{D}_l$ into the graph. We explain these steps below. An illustrative example is presented in Fig.~\ref{fig:semiag}. The workflow of SemiAG operations is presented in Fig.~\ref{fig:main} (c). \par

\noindent \textbf{Consensus KNN graph.}
Given an embedding graph $\mathbf{G}_{\mathcal{H}}=(\mathcal{V}, \mathcal{E})$ whose node set $\mathcal{V}=\{\mathbf{h}_i\}_{i = 1}^{N_G}$ contains $N_G$ embeddings and edge set is $\mathcal{E}=\{e_{i,j}= \mathbf{h}_i \cdot \mathbf{h}_j\}_{i,j = 1}^{N_G}$, we build a consensus graph $\mathbf{G}_c=(g_{i,j})_{i,j = 1}^{N_G}$ on $\mathcal{V}$ via consensus statistics. 
Each edge $g_{i,j}$ of $\mathbf{G}_c$ is defined as:
\begin{equation}
    g_{i,j}=\left\{
\begin{aligned}
&|\{\mathbf{h}_c| \mathbf{h}_i, \mathbf{h}_j \in \mathcal{O}_K(\mathbf{h}_c)), \forall \mathbf{h}_c \in \mathcal{V} \}| & i\neq j \\
&0 & i=j ,
\label{eq:consensus}
\end{aligned}
\right.
\end{equation}
where $\mathcal{O}_{K}(\mathbf{h}_c) = \texttt{argtopK}_{\mathbf{h}_j}(\{ \mathbf{h}_j \cdot \mathbf{h}_c | \mathbf{h}_j \in \mathcal{V} \})$ denotes the $K$-neighborhood of $\mathbf{h}_c \in \mathcal{V}$.
Then, we convert it into $\mathbf{\tilde{G}}_c$ by row normalization. 
However, consensus graph has a defect: the neighborhood consensus condition is rigorous and only considers local information, which means abundant potential positives are still unretrieved. \par 

\noindent \textbf{Affinity propagation with SemiPriori.}
To overcome this issue, we leverage the graph diffusion algorithm~\cite{tpg} on the probabilistic matrix $\mathbf{\tilde{G}}_c$ to propagate local affinities along multi-hop paths to characterize higher-order structural information and avoid degenerated solutions.
Specifically, we apply TPG diffusion algorithm~\cite{tpg}, which iteratively computes the diffused graph $\mathbf{\tilde{G}}_d$ as:
\begin{equation}
    \mathbf{\tilde{G}}_d^{(t+1)} = \mathbf{\tilde{G}}_c \mathbf{\tilde{G}}_d^{(t)} \mathbf{\tilde{G}}_c^T + \mathbf{I}, t=1, ..., \eta
    \label{eq:diffuse}
\end{equation}
where $\mathbf{I}$ is an identity matrix, and $\eta$ is the total diffusion step. $\mathbf{\tilde{G}}_d^{(t)}$ denotes the $t$-th step diffused graph and $\mathbf{\tilde{G}}_d^{(0)}=\mathbf{\tilde{G}}_c$. We denote the final diffused graph as $\mathbf{\tilde{G}}_d$. In Appendix~A, we provide more detailed descriptions. \par

However, the consensus graph and affinity propagation neglect abundant prior information in the labeled data. 
To address the issue, we incorporate SemiPriori, \ie, add sample-wise class labels as pairwise constraints to $\mathbf{\tilde{G}}_d$.
We set the edge to $1$ if two nodes have the same labels (\ie, $y_i = y_j$) and prune the edge if $y_i \neq y_j$. 
Meanwhile, we sparsify $\mathbf{\tilde{G}}_d$ with a pre-defined quantile $q$, then the generated binarized affinity graph $\mathbf{G}_b$ is denoted as:
\begin{equation}
    \mathbf{G}_b(i,j) = \left\{ \\
    \begin{aligned}
    & 1 & & (y_i = y_j) \lor \Big(\mathbf{\tilde{G}}_d(i,j)>q\Big) \\
    & 0 & & (y_i \neq y_j) \\
    \end{aligned}
    \right.
    \label{eq:semi-ag}
\end{equation}
On binarized affinity graph $\mathbf{G}_b$, positive/negative pairs are regarded as reliable pseudo positives/negatives in noisy embedding space for further contrastive affinity learning (in Sec.~\ref{method:cal}). 
Therefore, pseudo-labels of both labeled and unlabeled data are computed; while, those of labeled data are calibrated by SemiPriori.
Note that we compute two binarized graphs for {\tt [CLS]} and {\tt [P]} embeddings, respectively. \par

\subsection{Contrastive Affinity Learning Phase}
\label{method:cal}
In this section, given reliable pseudo positives identified from an embedding graph, we introduce two critical components for the second phase learning: online graph sampling strategy and our proposed CAL loss.
The overall framework of contrastive affinity learning is illustrated in Fig.~\ref{fig:main} (b).
\par 

\noindent \textbf{Graph sampling with memory.}
One practical issue arises (in Sec.~\ref{method:semi-ap}): SemiAG on mini-batches is not effective due to sampling insufficiency; while conducting SemiAG offline on the entire dataset is time-consuming and memory inefficiency~\cite{semi-prop}. 
To strike a balance between the graph size and computation resources, inspired by \cite{graph-sampling}, we dynamically construct a sub-graph $\mathbf{G}_{\mathcal{H}}^\prime$ sub-sampled from the entire graph $\mathbf{G}_{\mathcal{H}}$ supported by an extra embedding memory bank $\mathcal{M}$ and an exponentially moving averaged (EMA) teacher ($f_{\text{T}}, g_{\text{T}}$), like MoCo~\cite{moco}. 
Specifically, for each input batch, the EMA teacher produces stable embeddings, which are enqueued to the fixed-size first-in-first-out memory. 
The sub-graph $\mathbf{G}_{\mathcal{H}}^\prime$ is then constructed by the embeddings in the memory and teacher embeddings in the current batch.
We denote its node set as $\mathcal{V}(\mathbf{G}_{\mathcal{H}}^\prime) = \mathcal{M}\bigcup \{\mathbf{h}_{\text{T}}=f_{\text{T}}(\mathbf{x})| \mathbf{x}\in \mathcal{B}\}$. 
In this way, we can apply the same SemiAG operation (in Sec.~\ref{method:semi-ap}) to the sub-graph on the fly with adjustable memory sizes.
Note that we maintain another memory for SemiAG on prompts, since we retain DPR loss in contrastive affinity learning phase. \par

\noindent \textbf{Contrastive affinity loss.} 
The target of CAL loss is to gradually calibrate the semantic representation by learning from generated affinity constraints in graphs.
Given the sub-graph $\mathbf{G}_{\mathcal{H}}^\prime$ and its corresponding binarized graph $\mathbf{G}_b^\prime$ by SemiAG (in Sec.~\ref{method:semi-ap}), we formulate CAL loss with {\tt [CLS]} embedding $\mathbf{h}_i$ as a query, embeddings in sub-graph node set $\mathcal{V}(\mathbf{G}_{\mathcal{H}}^\prime)$ as keys:
\begin{equation}
    \begin{aligned}
        L_{\text{CAL}}^{\text{CLS}}(\mathbf{h}_i, \mathbf{G}_b^\prime) = L_{\text{con}}(\mathbf{h}_i, \mathcal{V}(\mathbf{G}_{\mathcal{H}}^\prime), \tau_a, \mathcal{P}_{a}, \mathcal{A}_{a}) \label{eq:cal}
    \end{aligned}
\end{equation}
where $\tau_a$ is a hyper-parameter, and the positive set is defined as $\mathcal{P}_{a}(\mathbf{h}_i)=\{\mathbf{h}_{\text{T}, j} | \mathbf{G}_b^\prime(i,j)=1, \forall \mathbf{h}_{\text{T}, j \neq i} \in \mathcal{V}(\mathbf{G}_{\mathcal{H}}^\prime)\}\cup \{\mathbf{h}_{\text{T}, i}^\prime\}$ where $\mathbf{h}_{\text{T}, i}^\prime$ is $\mathbf{h}_i$ augmented counterpart.
Note that $\mathcal{P}_{a}$ is always non-empty.
Since the whole $\mathcal{V}(\mathbf{G}_b^\prime)$ is too large, we define the anchor set $\mathcal{A}_a(\mathbf{h}_i)$ as the union of $\mathcal{P}_{a}(\mathbf{h}_i)$ and $N_{\text{neg}}$ randomly sampled pseudo-negatives for each query. 
For $L_{\text{CAL}}^{\text{CLS}}$ loss, we also define its corresponding DPR counterpart of CAL loss as $L_{\text{CAL}}^{\text{P}}$. \par

\noindent \textbf{Overall optimization objective.}
At CAL stage, we also preserve SemiCL loss in feature space to retain the model capability of instance-wise discrimination. 
To further increase the consistency between the teacher and student, we adapt supervised and self-supervised term of SemiCL (Eq.~\ref{eq:semi-cl}) as:
\begin{equation}
    \begin{aligned}
    L_{\text{self}}^{\text{CLS}}(\mathbf{z}) & = L_{\text{con}}\Big(\mathbf{z}, \mathbf{Z}_{\mathcal{B}, T}; \tau, \mathcal{P}_{\text{self}}, \mathcal{A}_{\text{self}} \Big) \\
    L_{\text{sup}}^{\text{CLS}}(\mathbf{z}) & = L_{\text{con}}\Big(\mathbf{z}, \mathbf{Z}_{\mathcal{B}_l, T}; \tau_a, \mathcal{P}_{\text{sup}}, \mathcal{A}_{\text{sup}}\Big) \mathbb{I}\Big(\mathbf{z} \in \mathbf{Z}_{\mathcal{B}_l}\Big)
    \end{aligned}
    \label{eq:semicl-2}
\end{equation}
Here, we use student feature $\mathbf{z}$ as a query and teacher features $\mathbf{Z}_{\mathcal{B}, T}, \mathbf{Z}_{\mathcal{B}_l, T}$ as keys to strengthen consistencies. 
The positive and anchor sets follow the same definition as in Eq.~(\ref{eq:semi-cl}) but are defined in the teacher feature space. \par

Then, the overall loss for {\tt [CLS]} token at CAL stage is formulated as:
\begin{equation}
    L_2^{\text{CLS}} = (1 - \alpha)L_{\text{sup}}^{\text{CLS}} + \alpha \Big(\beta L_{\text{CAL}}^{\text{CLS}} + (1 - \beta )L_{\text{self}}^{\text{CLS}} \Big)
    \label{eq:total-1}
\end{equation}
where $\beta$ is an adjustable weight. Its corresponding DPR counterpart can be similarly defined, denoted as $L_2^{\text{P}}$. \par

Finally, since we also adopt DPR at CAL stage, the overall optimization objective is formulated as:
\begin{equation}
    L_2 = L_2^{\text{CLS}} + \gamma L_2^{\text{P}}
    \label{eq:total-2}
\end{equation}
During the inference, the {\tt [CLS]} embeddings are adopted as final predictions. \par

\section{Experiments}
\subsection{Datasets}
\label{exp:dataset}
We evaluate \mbox{PromptCAL} on three generic datasets (\ie, CIFAR-10/100~\cite{cifar} and ImageNet-100~\cite{imagenet}) and three fine-grained datasets (\ie, CUB-200~\cite{dataset-cub}, StandfordCars~\cite{scars}, and Aircraft~\cite{aircraft}).
A summary of datasets is listed in Appendix B.
For each dataset, we first subsample $|\mathcal{C}_{kwn}|$ known classes from all classes. Then, a pre-defined ratio of images for known classes are sampled to form the labeled set $\mathcal{D}_l$. Follow GCD~\cite{gcd}, we set labeling ratio to $80\%$ for CIFAR-100 and $50\%$ for other datasets unless otherwise specified.
All unsampled images constitute $\mathcal{D}_u$.
In practice, we adopt the same dataset split of $\mathcal{D}_l$ and $\mathcal{D}_u$ as in~\cite{gcd}. (See Table~6 in Appendix~B for more details on known class numbers and labeling ratios for all dataset).
Besides, we adopt fewer $|\mathcal{C}_{kwn}|$ and smaller labeling ratios in more challenging setups for ablation study (Sec.~\ref{tab:challenge}). \par

\subsection{Evaluation Protocol}
\label{sec: settings}
We follow GCD~\cite{gcd} evaluation protocol in all experiments unless otherwise specified. 
Specifically, we perform SemiKMeans clustering~\cite{gcd} on the predicted embeddings.
Then, all clusters are mapped through the optimal assignment solved by Hungarian algorithm~\cite{hungarian} to their ground-truth classes.
The accuracy scores for {\tt All}, {\tt Known}, and {\tt New} classes are reported.
The predicted embeddings from the student class token are evaluated during inference. 

\subsection{Implementation Details}
\label{sec: imp}
Following GCD~\cite{gcd}, we use ViT-B/16 pre-trained DINO~\cite{DINO} on ImageNet-1K~\cite{imagenet} as our backbone for evaluation. 
For all experiments, we fix the batch size to $128$ and use the same data augmentation strategies as~\cite{gcd}. 
We present complete implementation details in Appendix C. \par

\subsection{Main Results}
\label{exp:main}
\noindent \textbf{Evaluation on generic datasets.}
We evaluate both stages of \mbox{PromptCAL} on three generic datasets (\ie, CIFAR-10/100~\cite{cifar}, and ImageNet-100~\cite{imagenet}).
Table \ref{tab:main-gen} shows that our \mbox{PromptCAL} consistently and significantly surpasses previous SoTAs, \ie, ViT-adapted ORCA~\cite{orca}, our baseline GCD~\cite{gcd}, and adapted NCD SOTA methods (UNO+~\cite{uno} and RankStats+~\cite{rs}) in terms of overall accuracy on all three datasets.
Specifically, PromptCAL surpasses GCD by $6.4\%$ on CIFAR-10, $8.2\%$ on CIFAR-100, and $9.0\%$ ImageNet-100 on {\tt All} classes; it also remarkably outperforms ORCA by $7\%$ on CIFAR-100 and $3.9\%$ on ImageNet-100.
Besides, in contrast to ORCA and UNO+ which suffer from severe overfitting on {\tt Known} classes, \mbox{PromptCAL} manifests substantial advantages over other methods on {\tt New} classes (about $10\%$ improvements on three datasets). \par 

By comparing the $1^{st}$ stage (\mbox{PromptCAL}-$1^{st}$) with the $2^{nd}$ stage (\mbox{PromptCAL}-$2^{nd}$), we observe major performance boosts, especially on {\tt New} classes.
In addition, we also notice that both stages of our \mbox{PromptCAL} have significant contributions to the final performance on generic datasets.
Specifically, \mbox{PromptCAL-$1^{st}$} improves $5.6\%$ and $3.0\%$ over GCD on CIFAR-10/100, respectively; while the \mbox{PromptCAL-$2^{nd}$} further improves by $5.2\%$ and $9.0\%$ on CIFAR-100 and ImageNet-100, respectively. Besides, we also achieve $\sim7\%$ boost of overall accuracy on CIFAR-100 and $4\%$ on ImageNet-100 when compared with ORCA.
Therefore, above results validate advantages and effectiveness of our two-stage \mbox{PromptCAL} in category discovery. \par

\begin{table}[]
    \centering
    \resizebox{\linewidth}{!}{
    \begin{tabular}{c||ccc|ccc|ccc}
    \toprule
         & \multicolumn{3}{c|}{\textbf{CIFAR-10}} & \multicolumn{3}{c|}{\textbf{CIFAR-100}} & \multicolumn{3}{c}{\textbf{ImageNet-100}} \\
        \textbf{Method} & All & Known & New & All & Known & New & All & Known & New \\
    \midrule
        KMeans~\cite{kmeanspp} & 83.6 & 85.7 & 82.5 & 52.0 & 52.2 & 50.8 & 72.7 & 75.5 & 71.3 \\
        RankStats+~\cite{rs} &  46.8 & 19.2 & 60.5 & 58.2 & 77.6 & 19.3 & 37.1 & 61.6 & 24.8 \\
        UNO+~\cite{uno} & 68.6 & \textbf{98.3} & 53.8 & 69.5 & 80.6 & 47.2 & 70.3 & \textbf{95.0} & 57.9 \\
        GCD~\cite{gcd} & 91.5 & 97.9 & 88.2 & 73.0 & 76.2 & 66.5 & 74.1 & 89.8 & 66.3 \\
        ORCA$^\dag$~\cite{orca} & 96.9 & 95.1 & 97.8 & 74.2 & 82.1 & 67.2 & 79.2 & 93.2 & 72.1 \\
        \midrule
        \rowcolor{Green}
        \textbf{PromptCAL-$1^{st}$ (Ours)} & 97.1 & 97.7 & 96.7 & 76.0 & 80.8 & 66.6 & 75.4 & 94.2 & 66.0 \\
        \rowcolor{Green}
        \textbf{PromptCAL-$2^{nd}$ (Ours)} & \textbf{97.9} & 96.6 & \textbf{98.5} & \textbf{81.2} & \textbf{84.2} & \textbf{75.3} & \textbf{83.1} & 92.7 & \textbf{78.3} \\
     \bottomrule
    \end{tabular}
    }
    \vspace{-5pt}
    \caption{\textbf{Evaluation on three generic datasets.}  Accuracy scores are reported. \dag denotes adapted methods. Both stages of PromptCAL are evaluated.}
    \vspace{-10pt}
    \label{tab:main-gen}
\end{table}

\begin{table}[]
    \centering
    \resizebox{\linewidth}{!}{
    \begin{tabular}{c||ccc|ccc|ccc}
    \toprule
         & \multicolumn{3}{c|}{\textbf{CUB-200}} & \multicolumn{3}{c|}{\textbf{StanfordCars}} & \multicolumn{3}{c}{\textbf{Aircraft}} \\
        \textbf{Method} & All & Known & New & All & Known & New & All & Known & New \\
    \midrule
        KMeans~\cite{kmeanspp} & 34.3 & 38.9 & 32.1 & 12.8 & 10.6 & 13.8 & 12.9 & 12.9 & 12.8 \\
        RankStats+~\cite{rs} & 33.3 & 51.6 & 24.2 & 28.3 & 61.8 & 12.1 & 27.9 & \textbf{55.8} & 12.8 \\
        UNO+~\cite{uno} & 35.1 & 49.0 & 28.1 & 35.5 & \textbf{70.5} & 18.6 & 28.3 & 53.7 & 14.7 \\
        GCD~\cite{gcd} & 51.3 & 56.6 & 48.7 & 39.0 & 57.6 & 29.9 & 45.0 & 41.1 & 46.9 \\
        ORCA$^\dag$~\cite{orca} & 36.3 & 43.8 & 32.6 & 31.9 & 42.2 & 26.9 & 31.6 & 32.0 & 31.4 \\
        \midrule
        \rowcolor{Green}
        \textbf{PromptCAL-$1^{st}$ (Ours)} & 51.1 & 55.4 & 48.9 & 42.6 & 62.8 & 32.9 & 44.5 & 44.6 & 44.5 \\
        \rowcolor{Green}
        \textbf{PromptCAL-$2^{nd}$ (Ours)} & \textbf{62.9} & \textbf{64.4} & \textbf{62.1} & \textbf{50.2} & 70.1 & \textbf{40.6} & \textbf{52.2} & 52.2 & \textbf{52.3} \\
     \bottomrule
    \end{tabular}
    }
    \vspace{-5pt}
    \caption{\textbf{Evaluation on three fine-grained datasets.} Accuracy scores are reported. \dag denotes adapted methods. Both stages of PromptCAL are evaluated.}
    \vspace{-10pt}
    \label{tab:main-fg}
\end{table}

\noindent \textbf{Evaluation on fine-grained datasets.}
We also report results on fine-grained datasets to demonstrate the \mbox{PromptCAL} effectiveness in Table \ref{tab:main-fg}.
Apparently, the low performance of KMeans illustrates the challenging nature of fine-grained category discovery caused by larger intra-class and lower inter-class variations.
Notice that ORCA performance degrades substantially on three fine-grained datasets.
In contrast, our \mbox{PromptCAL} consistently exceeds NCD SOTA and ORCA, and outperforms GCD by $\sim$~$11\%$ on {\tt All} classes on CUB-200 and StanfordCars and $\sim$~$7\%$ on Aircraft. Different from results in Table \ref{tab:main-gen}, the results on fine-grained datasets show that the major performance gain of \mbox{PromptCAL} originates from the $2^{nd}$ CAL stage.
Noticeably, \mbox{PromptCAL-$1^{st}$} performance even drops compared with GCD on CUB-200 and Aircraft datasets; while, \mbox{PromptCAL-$2^{nd}$} achieves remarkable and consistent improvements, especially on {\tt New} classes. \par

\subsection{Ablation and analysis}
\label{exp:ablation}
In this section, we conduct extensive ablation experiments to reveal and investigate contributions of each component. 
Next, we present in-depth analysis on the effectiveness of SemiAG and discuss the effect of visual prompts in \mbox{PromptCAL}.
Further, we explore how \mbox{PromptCAL} performs in more challenging and real-world scenarios with lower-labeling and fewer-classes.
Finally, we present additional ablation results in Appendix D, and additional qualitative results in Appendix F. \par

\begin{table}[]
    \centering
    \resizebox{\linewidth}{!}{
    \begin{tabular}{ccccc|ccc}
    \toprule
         & \textbf{cKNN} & \textbf{AP} & \textbf{SemiPriori} & \textbf{SemiCL} & All & Known & New \\
    \midrule
        (1) & \XSolidBrush & \XSolidBrush & \Checkmark & \Checkmark & \redscore{60.1}{-2.8} & \greenscore{70.1}{+5.7} & \redscore{55.1}{-7.0} \\
        (2) & \Checkmark & \Checkmark & \Checkmark & \XSolidBrush & \redscore{61.7}{-1.2} & \redscore{63.6}{-0.8} & \redscore{60.7}{-1.4} \\
        (3) & \Checkmark & \Checkmark & \XSolidBrush & \Checkmark & \redscore{57.3}{-5.6} & \redscore{61.8}{-2.6} & \redscore{55.1}{-7.0} \\
        (4) & \Checkmark & \XSolidBrush & \Checkmark & \Checkmark & \redscore{54.6}{-8.3} & \greenscore{65.5}{+1.1} & \redscore{49.1}{-13.0} \\
        \rowcolor{Green}
        (5) & \Checkmark & \Checkmark & \Checkmark & \Checkmark & 62.9 & 64.4 & 62.1 \\
    \bottomrule
    \end{tabular}
    }
    \caption{\textbf{Ablation study on effectiveness of SemiAG in CAL stage on CUB-200~\cite{dataset-cub} dataset.} Here, {\bf cKNN:} consensus KNN graph; {\bf AP:} affinity propagation; {\bf SemiPriori:} semi-supervised prior knowledge; {\bf SemiCL:} semi-supervised contrastive loss in projected feature space on {\tt [CLS]} and {\tt [P]}. Scores reported in clustering accuracy. 
    Each proposed component favorably contributes to the overall performance.}
    \vspace{-10pt}
    \label{tab:ablation-cal}
\end{table}

\begin{table}[]
    \centering
    \resizebox{\linewidth}{!}{
    \begin{tabular}{cccc}
        \includegraphics[scale=0.15]{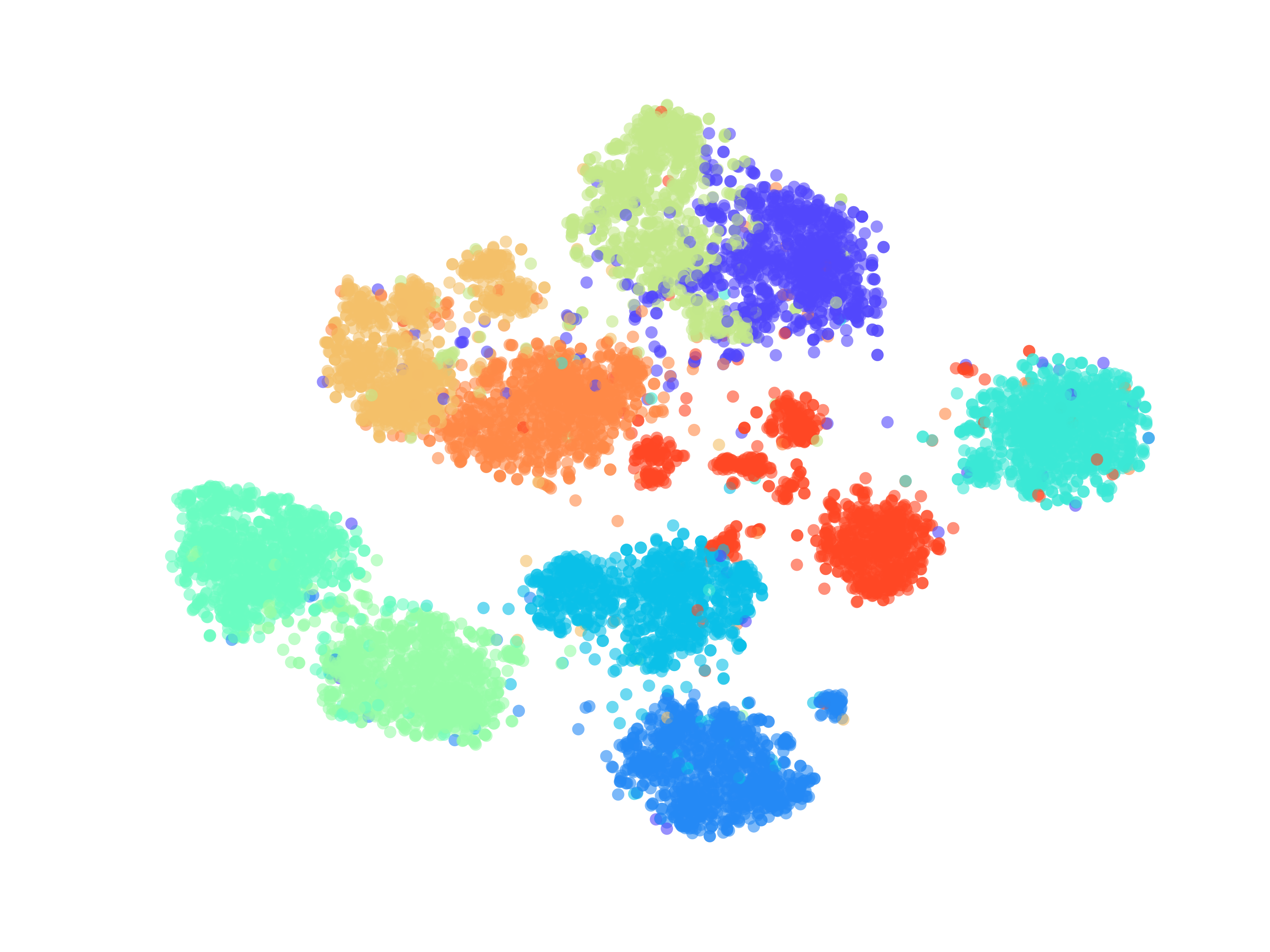} & \includegraphics[scale=0.15]{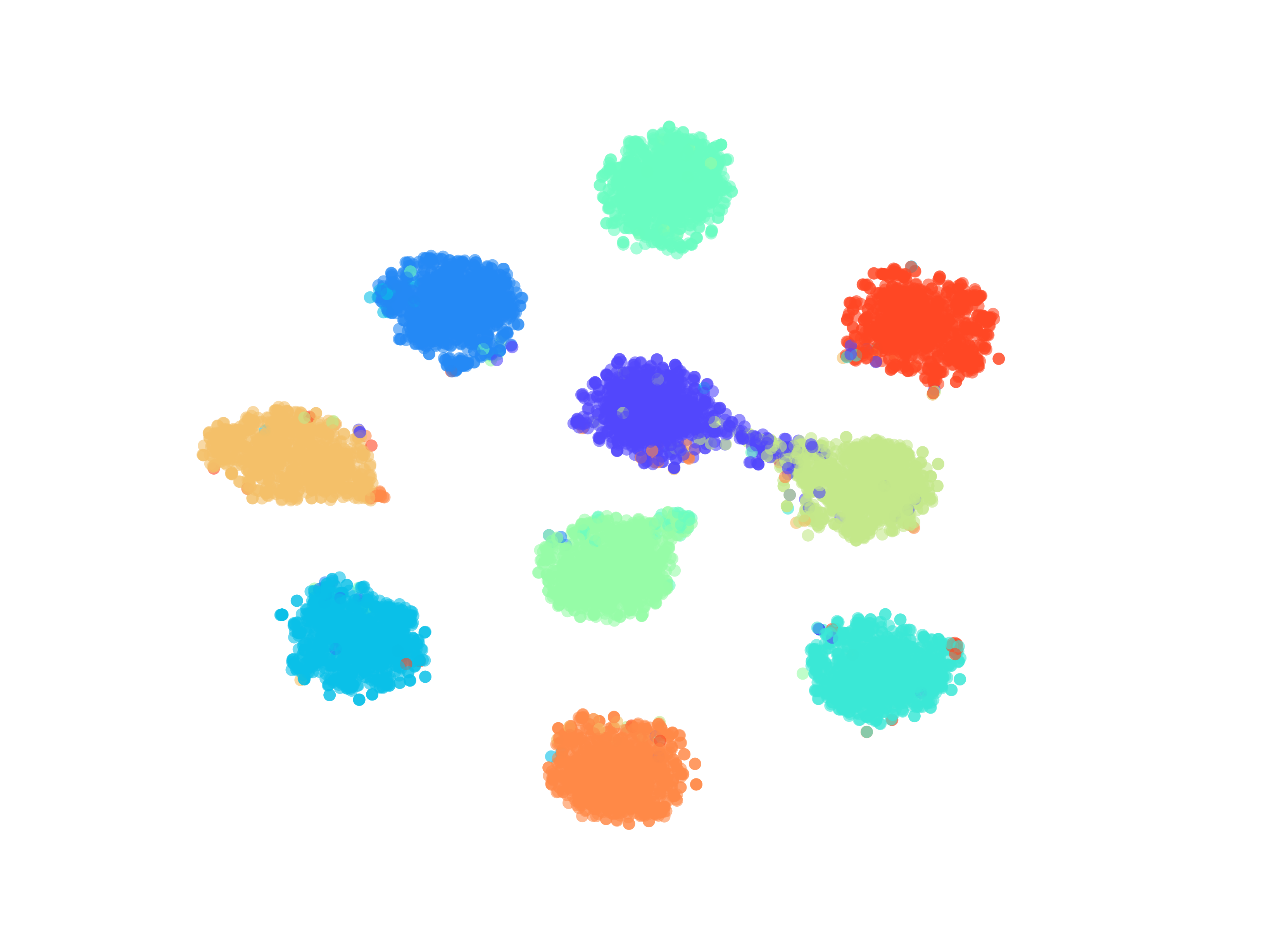} &         \includegraphics[scale=0.15]{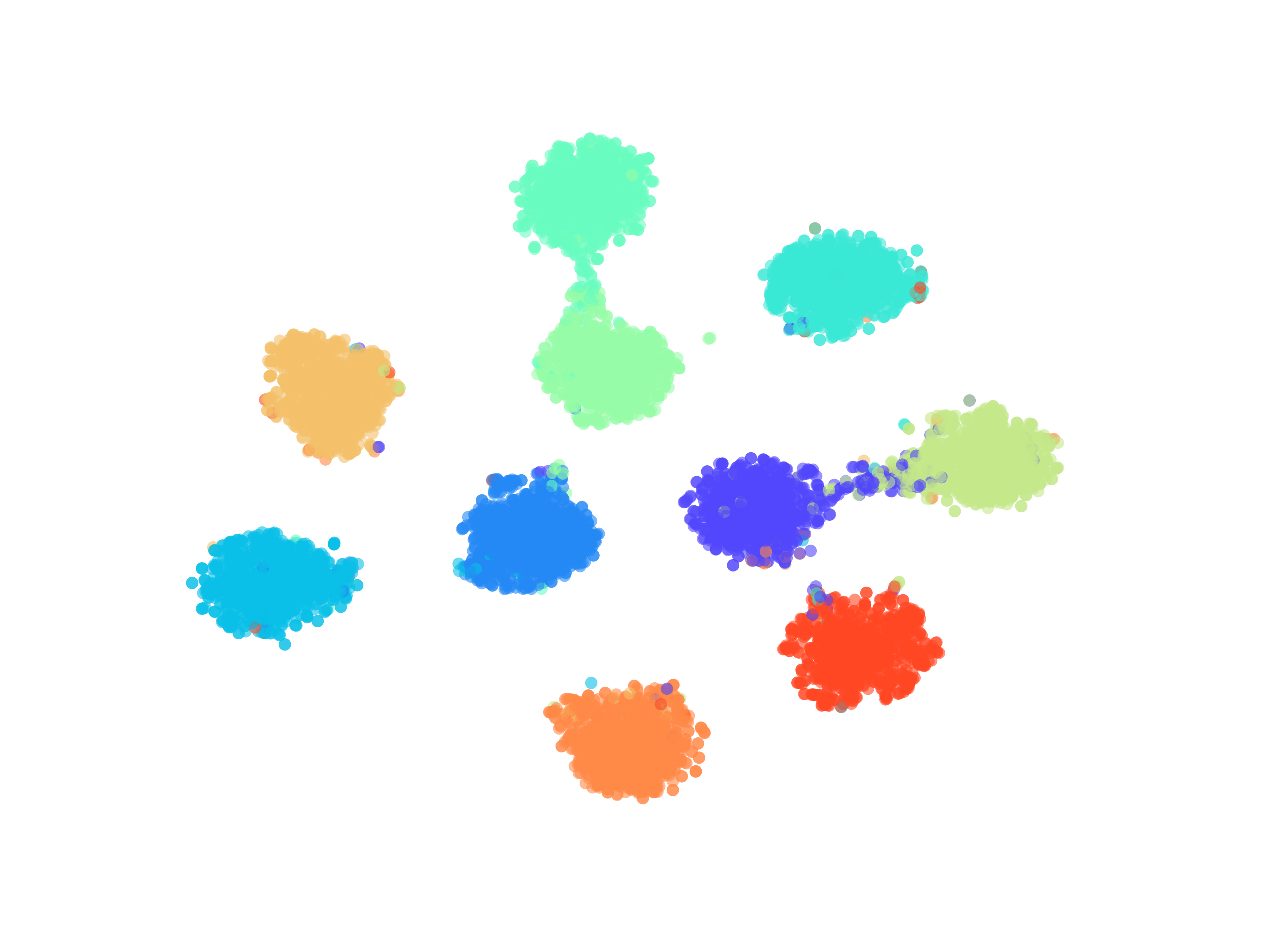} &
        \includegraphics[scale=0.15]{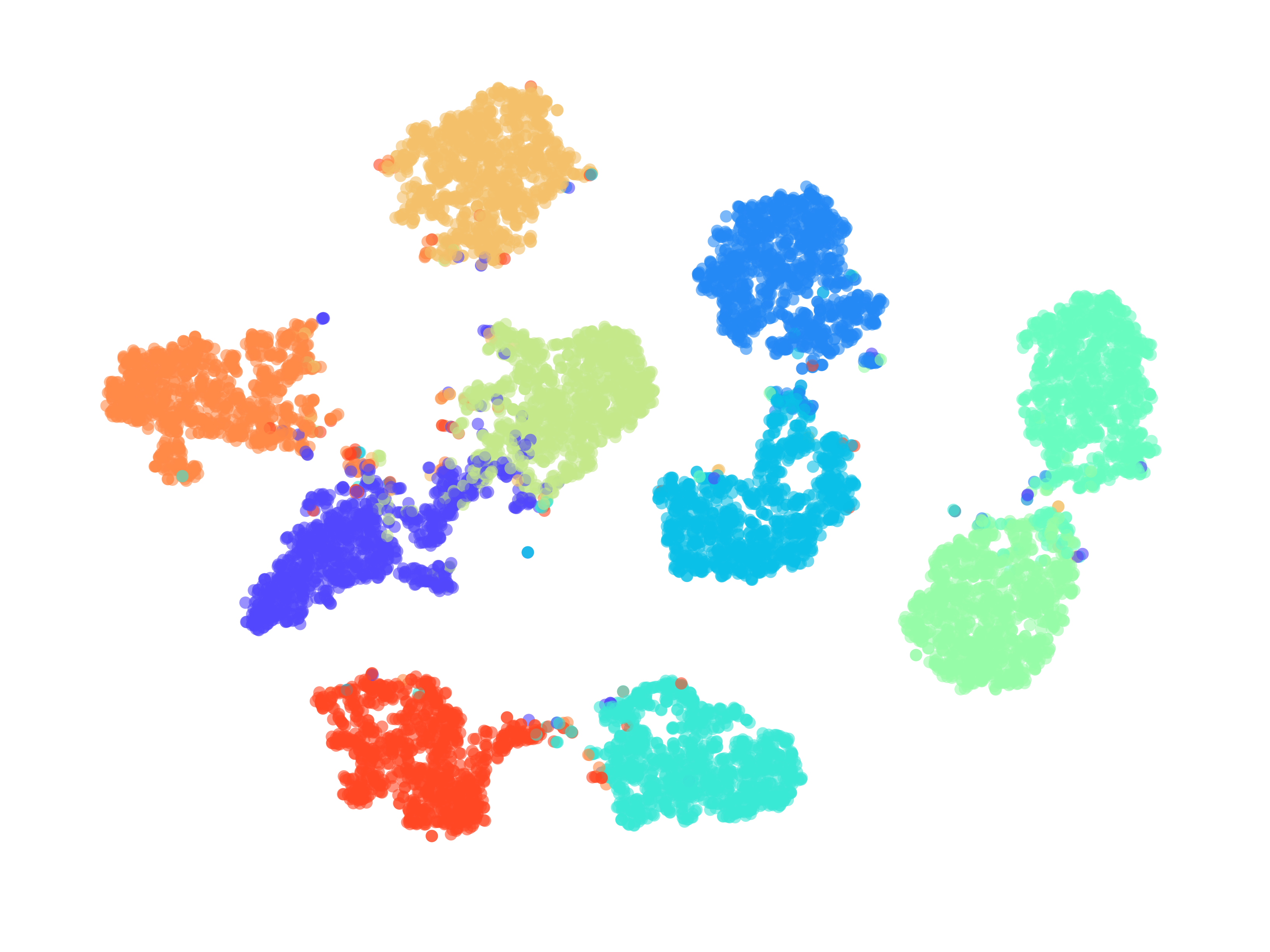} \\
        (a) & (b) & (c) & (d) \\
    \end{tabular}
    }
    \vspace{-5pt}
    \captionof{figure}{\textbf{The t-SNE~\cite{tsne} visualization of ViT embeddings on CIFAR-10 test set}. (a) is {\tt [CLS]} embeddings from naive VPT model; (b) denotes our PromptCAL {\tt [CLS]} embeddings; (c) denotes our PromptCAL ensembled {\tt [P]} embeddings; (d) represents embeddings of an arbitraty PromptCAL unsupervised prompt.
    All figures share the same axis scale. The complete visualization is presented in Appendix D.}
    \label{tab:tsne-1}
    \vspace{-15pt}
\end{table}

\noindent \textbf{Effectiveness of contrastive affinity learning.}
As mentioned in Sec.~\ref{exp:main}, SemiAG dominates the large improvements of \mbox{PromptCAL}.
First, we conduct ablation experiments on SemiAG in CAL stage, in Table \ref{tab:ablation-cal}. The $1^{st}$ row denotes the performance of using naive KNN with SemiPriori for pseudo labeling at CAL stage; while, the last row represents our full SemiAG setup.
The $2^{nd}$, $3^{rd}$, and $4^{th}$ row represent \mbox{PromptCAL} without semi-supervised contrastive loss, semi-supervised prior knowledge (Sec.~\ref{method:semi-ap}), and affinity propagation (Sec.~\ref{method:semi-ap}), respectively. 
From the results, we can observe that incorporating each component has a clear contribution: 
{\tt (a)} Naive KNN severely overfits {\tt Known} and performs poorer (with nearly $2.8\%$ and $7.0\%$ accuracy drops on {\tt All} and {\tt New} classes, respectively) than SemiAG, due to its susceptibility to noisy neighborhoods.
{\tt (b)} Affinity propagation is the most consequential component (improving by $8.3\%$ on {\tt All} and $13\%$ on {\tt New}), which proves the importance of counteracting adverse effects of false negatives in contrastive loss by retrieving more reliable positives.
{\tt (c)} Retaining SemiCL is beneficial, which, we guess, is because it can push away noisy pseudo positive samples and, thus, prevent overfitting and degenerated solutions.
{\tt (d)} SemiPriori further benefits overall performance by about $5.6\%$ on {\tt All} and $7\%$ on {\tt New}, which manifests the importance of incorporating the prior knowledge to calibrate pseudo labels. \par

\begin{table}[]
    \centering
    \resizebox{\linewidth}{!}{
    \begin{tabular}{ccccc|ccc}
    \toprule
         & \textbf{Prompt} & $\mathbf{L_{\text{semi}}^{\text{P}}}$ & $\mathbf{L_{\text{CAL}}^{\text{P}}}$ & \textbf{CAL stage} & All & Known & New \\
    \midrule
        (1) & \XSolidBrush & \XSolidBrush & \XSolidBrush & \XSolidBrush & \redscore{51.3}{-11.6} & \redscore{56.6}{-7.8} & \redscore{48.7}{-12.4}  \\
        (2) & \Checkmark & \XSolidBrush & \XSolidBrush & \XSolidBrush & \redscore{51.1}{-11.8} & \redscore{55.4}{-9.0} & \redscore{48.9}{-12.2}  \\
        (3) & \Checkmark & \Checkmark & \XSolidBrush & \Checkmark & \redscore{61.6}{-1.3} & \greenscore{68.9}{+4.5} & \redscore{58.0}{-4.1} \\
        (4) & \Checkmark & \XSolidBrush & \Checkmark & \Checkmark & \redscore{61.2}{-1.7} & \greenscore{65.2}{+0.8} & \redscore{59.2}{-2.9} \\
        (5) & \XSolidBrush & \XSolidBrush & \XSolidBrush & \Checkmark & \redscore{60.3}{-2.6} & \greenscore{64.8}{+0.4} & \redscore{58.0}{-4.1} \\
        \rowcolor{Green}
        (6) & \Checkmark & \Checkmark & \Checkmark & \Checkmark & 62.9 & 64.4 & 62.1 \\
    \bottomrule
    \end{tabular}
    }
    \vspace{-5pt}
    \caption{\textbf{Ablation study on effectiveness of prompt-related components on CUB-200 dataset.}
    Here, {\bf Prompt:} prompt-adapted backbone; {\bf $L_{\text{semi}}^{\text{P}}$:} semi-supervised contrastive loss on {\tt [P]} prompts; {\bf $L_{\text{CAL}}^{\text{P}}$:} CAL loss on {\tt [P]}; {\bf CAL stage:} second-stage training.
    Scores reported in clustering accuracy. 
    Each component favorably contributes to the overall performance gain.
    }
    \vspace{-5pt}
    \label{tab:ablation-prompt}
\end{table}

\begin{table}[t]
    \centering
    \resizebox{\linewidth}{!}{
    \begin{tabular}{c||ccc|ccc|ccc}
    \toprule
         & \multicolumn{3}{c|}{C50-L10} & \multicolumn{3}{c|}{C25-L50} & \multicolumn{3}{c}{C10-L50} \\
     \midrule
       \textbf{Method} & All & Known & New & All & Known & New & All & Known & New \\
    \midrule
        GCD~\cite{gcd} & 60.2 & 68.9 & 55.8 & 56.8 & 67.6 & 55.0 & 48.3 & 65.1 & 47.3 \\
        ORCA (ResNet)~\cite{orca} & 39.4 & 55.1 & 31.2 & 37.0 & 64.1 & 31.7 & 30.1 & 64.3 & 27.1 \\
        ORCA$^\dag$ (ViT)~\cite{orca} & 60.3 & 66.0 & 55.3 & 58.2 & \textbf{79.9} & 57.5 & 51.7 & 78.0 & 50.2 \\
        \midrule 
        \rowcolor{Green}
        \textbf{PromptCAL-$1^{st}$ (Ours)} & 62.7 & 74.7 & 56.6 & 60.2 & 70.7 & 58.5 & 48.7 & 68.4 & 47.6 \\
        \rowcolor{Green}
        \textbf{PromptCAL-$2^{nd}$ (Ours)} & \textbf{68.9} & \textbf{77.5} & \textbf{64.7} & \textbf{65.7} & 76.9 & \textbf{63.8} & \textbf{53.2} & \textbf{79.3} & \textbf{51.7} \\
     \bottomrule
    \end{tabular}
    }
    \vspace{-5pt}
    \caption{\textbf{Ablation study on few-annotation GNCD on CIFAR-100~\cite{cifar} dataset.} Digits following 'C' and 'L' stand for percentages of known classes and labeling ratios. \dag denotes adapted methods. Scores reported in accuracy.}
    \vspace{-10pt}
    \label{tab:challenge}
\end{table}

\noindent \textbf{Role of discriminative prompt regularization.}
Table \ref{tab:ablation-prompt} presents the ablation results for prompt-related components of \mbox{PromptCAL}. 
The $1^{st}$ and $2^{nd}$ rows denote the GCD baseline and our warmed-up \mbox{PromptCAL-$1^{st}$}. 
We note that visual prompts make no significant difference to the performance. However, we argue that it is due to lack of semantic discriminative supervision. 
Specifically, by observing \mbox{PromptCAL} without semantic discrimination supervision ($3^{rd}$ row) and \mbox{PromptCAL} without sample discrimination supervision ($4^{th}$ row), we can infer that both semantic discriminativeness and sample-wise discriminativeness are critical.
Generally, lack of semantic discriminativeness will cause severe overfitting on {\tt Known} classes.
Furthermore, semantic prompt tuning is beneficial for discovering novel classes, since \mbox{PromptCAL} surpasses its counterpart without any prompt-related component ($5^{th}$ row) on {\tt New} by $2.6\%$.
To summarize, semantically-aware DPR plays a positive and auxiliary role in facilitating semantic discriminativeness especially in categorizing novel classes. 
In fact, we conclude from additional ablations in Appendix~D that the gains of prompts are more significant on larger datasets. \par

To vividly illustrate this point, we present the t-SNE~\cite{tsne} visualization results in Fig.~\ref{tab:tsne-1} (see complete results in Appendix~D.3).
Here, we directly draw conclusions that {\tt (a)} naive VPT causes overclustering problem and lacks semantically discriminativeness; {\tt (b)} our proposed DPR supervision increases semantic discriminativeness of supervised and unsupervised prompts, which further enhances semantic signals of DPR loss and enables DPR and CAL to synergistically improve the overall performance.
We present more discussions on this in Appendix~D.3. \par

\noindent \textbf{Towards few-annotation GNCD.}
We further evaluate our \mbox{PromptCAL} against other SOTA methods on more challenging few-annotation setups on CIFAR-100 dataset, \ie, fewer known classes and lower labeling ratios.
We consider three setups in Table \ref{tab:challenge}: (1) C50-L10: $50\%$ classes are known in which $10\%$ samples are labeled; (2) C25-L50: $25\%$ classes are known in which $50\%$ samples are labeled; (3) C10-L50: $10\%$ classes are known in which $50\%$ samples are labeled.
Since the few-annotation can incur more open-set noises, we set $K=5$ for PromptCAL to increase robustness to noisy pseudo-labels. \par

From results in Table \ref{tab:challenge}, we conclude that \mbox{PromptCAL} is robust to both low-labeling and few-class scenarios, outcompeting all SoTAs with large margins.
Practically, it is more demanding for models to infer novel semantic clustering when fewer classes are known under semantic shifts. This explains the lower performance of all models in setup (3) than in setup (1).
Compared with GCD and ORCA, our \mbox{PromptCAL} can learn semantically robust representation and consistently achieve high performance in all setups.
ORCA (ViT)~\cite{orca} achieves stronger performance than GCD; while, our PromptCAL can still outperform ORCA with clear margins in all setups.
For example, \mbox{PromptCAL} surpasses ORCA (ViT) by $\sim$~$8\%$ on {\tt All} accuracy in C50-L10 and C25-L50.
We again observe that \mbox{PromptCAL}-$2^{nd}$ learning contributes most to the overall performance, which again proves our proposed method can effectively calibrate the learned representation with remarkable gains on {\tt New} classes. This capability best suits GNCD problem. \par

\section{Conclusion}
In this paper, we propose a two-stage framework, \mbox{PromptCAL}, to tackle challenging GNCD problem.
After the warm-up stage of semi-supervised contrastive learning, we iteratively and simultaneously conduct contrastive affinity learning and discriminative prompt regularization to calibrate semantic representations.
Specifically, at each iteration, we leverage discovered pseudo affinities on generated affinity graphs to guide optimization of the class token and to reinforce the semantic discriminativeness of prompts and our prompt-adapted ViT backbone.
Extensive experiments on multiple generic and fine-grained benchmarks showcase that \mbox{PromptCAL} achieves state-of-the-art performance.
Additional evidences illustrates that our discriminative prompt regularization and contrastive affinity learning objectives achieve a synergistic effect. 
Moreover, \mbox{PromptCAL} exhibits remarkable gains on few-class and low-label settings for categorizing novel classes. \par

{\small
\bibliographystyle{ieee_fullname}
\bibliography{egbib}

\begin{thebibliography}{10}\itemsep=-1pt

\bibitem{kmeanspp}
David Arthur and Sergei Vassilvitskii.
\newblock k-means++: The advantages of careful seeding.
\newblock Technical report, Stanford, 2006.

\bibitem{ckmeans}
Paul~S Bradley, Kristin~P Bennett, and Ayhan Demiriz.
\newblock Constrained k-means clustering.
\newblock {\em Microsoft Research, Redmond}, 20(0):0, 2000.

\bibitem{semi-vit}
Zhaowei Cai, Avinash Ravichandran, Paolo Favaro, Manchen Wang, Davide Modolo,
  Rahul Bhotika, Zhuowen Tu, and Stefano Soatto.
\newblock Semi-supervised vision transformers at scale.
\newblock {\em arXiv preprint arXiv:2208.05688}, 2022.

\bibitem{orca}
Kaidi Cao, Maria Brbic, and Jure Leskovec.
\newblock Open-world semi-supervised learning.
\newblock {\em arXiv preprint arXiv:2102.03526}, 2021.

\bibitem{deep-cluster}
Mathilde Caron, Piotr Bojanowski, Armand Joulin, and Matthijs Douze.
\newblock Deep clustering for unsupervised learning of visual features.
\newblock In {\em Proceedings of the European conference on computer vision
  (ECCV)}, pages 132--149, 2018.

\bibitem{DINO}
Mathilde Caron, Hugo Touvron, Ishan Misra, Herv{\'e} J{\'e}gou, Julien Mairal,
  Piotr Bojanowski, and Armand Joulin.
\newblock Emerging properties in self-supervised vision transformers.
\newblock In {\em Proceedings of the IEEE/CVF International Conference on
  Computer Vision}, pages 9650--9660, 2021.

\bibitem{ifnd}
Tsai-Shien Chen, Wei-Chih Hung, Hung-Yu Tseng, Shao-Yi Chien, and Ming-Hsuan
  Yang.
\newblock Incremental false negative detection for contrastive learning.
\newblock {\em arXiv preprint arXiv:2106.03719}, 2021.

\bibitem{little-help}
Debidatta Dwibedi, Yusuf Aytar, Jonathan Tompson, Pierre Sermanet, and Andrew
  Zisserman.
\newblock With a little help from my friends: Nearest-neighbor contrastive
  learning of visual representations.
\newblock In {\em Proceedings of the IEEE/CVF International Conference on
  Computer Vision}, pages 9588--9597, 2021.

\bibitem{uno}
Enrico Fini, Enver Sangineto, St{\'e}phane Lathuili{\`e}re, Zhun Zhong, Moin
  Nabi, and Elisa Ricci.
\newblock A unified objective for novel class discovery.
\newblock In {\em Proceedings of the IEEE/CVF International Conference on
  Computer Vision}, pages 9284--9292, 2021.

\bibitem{os-survey}
Chuanxing Geng, Sheng-jun Huang, and Songcan Chen.
\newblock Recent advances in open set recognition: A survey.
\newblock {\em IEEE transactions on pattern analysis and machine intelligence},
  43(10):3614--3631, 2020.

\bibitem{rs}
Kai Han, Sylvestre-Alvise Rebuffi, Sebastien Ehrhardt, Andrea Vedaldi, and
  Andrew Zisserman.
\newblock Automatically discovering and learning new visual categories with
  ranking statistics.
\newblock {\em arXiv preprint arXiv:2002.05714}, 2020.

\bibitem{dtc}
Kai Han, Andrea Vedaldi, and Andrew Zisserman.
\newblock Learning to discover novel visual categories via deep transfer
  clustering.
\newblock In {\em Proceedings of the IEEE/CVF International Conference on
  Computer Vision}, pages 8401--8409, 2019.

\bibitem{ptr}
Xu Han, Weilin Zhao, Ning Ding, Zhiyuan Liu, and Maosong Sun.
\newblock Ptr: Prompt tuning with rules for text classification.
\newblock {\em arXiv preprint arXiv:2105.11259}, 2021.

\bibitem{moco}
Kaiming He, Haoqi Fan, Yuxin Wu, Saining Xie, and Ross Girshick.
\newblock Momentum contrast for unsupervised visual representation learning.
\newblock In {\em Proceedings of the IEEE/CVF conference on computer vision and
  pattern recognition}, pages 9729--9738, 2020.

\bibitem{cnn-recognition}
Samer Hijazi, Rishi Kumar, Chris Rowen, et~al.
\newblock Using convolutional neural networks for image recognition.
\newblock {\em Cadence Design Systems Inc.: San Jose, CA, USA}, 9, 2015.

\bibitem{fnc}
Tri Huynh, Simon Kornblith, Matthew~R Walter, Michael Maire, and Maryam
  Khademi.
\newblock Boosting contrastive self-supervised learning with false negative
  cancellation.
\newblock In {\em Proceedings of the IEEE/CVF Winter Conference on Applications
  of Computer Vision}, pages 2785--2795, 2022.

\bibitem{semi-prop}
Ahmet Iscen, Giorgos Tolias, Yannis Avrithis, and Ondrej Chum.
\newblock Label propagation for deep semi-supervised learning.
\newblock In {\em Proceedings of the IEEE/CVF Conference on Computer Vision and
  Pattern Recognition}, pages 5070--5079, 2019.

\bibitem{vpt}
Menglin Jia, Luming Tang, Bor-Chun Chen, Claire Cardie, Serge Belongie, Bharath
  Hariharan, and Ser-Nam Lim.
\newblock Visual prompt tuning.
\newblock {\em arXiv preprint arXiv:2203.12119}, 2022.

\bibitem{wta}
Xuhui Jia, Kai Han, Yukun Zhu, and Bradley Green.
\newblock Joint representation learning and novel category discovery on
  single-and multi-modal data.
\newblock In {\em Proceedings of the IEEE/CVF International Conference on
  Computer Vision}, pages 610--619, 2021.

\bibitem{prompt-ensemble-1}
Zhengbao Jiang, Frank~F Xu, Jun Araki, and Graham Neubig.
\newblock How can we know what language models know?
\newblock {\em Transactions of the Association for Computational Linguistics},
  8:423--438, 2020.

\bibitem{slic}
Salar~Hosseini Khorasgani, Yuxuan Chen, and Florian Shkurti.
\newblock Slic: Self-supervised learning with iterative clustering for human
  action videos.
\newblock In {\em Proceedings of the IEEE/CVF Conference on Computer Vision and
  Pattern Recognition}, pages 16091--16101, 2022.

\bibitem{scl}
Prannay Khosla, Piotr Teterwak, Chen Wang, Aaron Sarna, Yonglong Tian, Phillip
  Isola, Aaron Maschinot, Ce Liu, and Dilip Krishnan.
\newblock Supervised contrastive learning.
\newblock {\em Advances in Neural Information Processing Systems},
  33:18661--18673, 2020.

\bibitem{sk}
Philip~A Knight.
\newblock The sinkhorn--knopp algorithm: convergence and applications.
\newblock {\em SIAM Journal on Matrix Analysis and Applications},
  30(1):261--275, 2008.

\bibitem{scars}
Jonathan Krause, Michael Stark, Jia Deng, and Li Fei-Fei.
\newblock 3d object representations for fine-grained categorization.
\newblock In {\em Proceedings of the IEEE international conference on computer
  vision workshops}, pages 554--561, 2013.

\bibitem{cifar}
Alex Krizhevsky, Geoffrey Hinton, et~al.
\newblock Learning multiple layers of features from tiny images.
\newblock 2009.

\bibitem{imagenet}
Alex Krizhevsky, Ilya Sutskever, and Geoffrey~E Hinton.
\newblock Imagenet classification with deep convolutional neural networks.
\newblock {\em Communications of the ACM}, 60(6):84--90, 2017.

\bibitem{dense-sup}
Chen-Yu Lee, Saining Xie, Patrick Gallagher, Zhengyou Zhang, and Zhuowen Tu.
\newblock Deeply-supervised nets.
\newblock In {\em Artificial intelligence and statistics}, pages 562--570.
  PMLR, 2015.

\bibitem{graph-sampling}
Jure Leskovec and Christos Faloutsos.
\newblock Sampling from large graphs.
\newblock In {\em Proceedings of the 12th ACM SIGKDD international conference
  on Knowledge discovery and data mining}, pages 631--636, 2006.

\bibitem{comatch}
Junnan Li, Caiming Xiong, and Steven~CH Hoi.
\newblock Comatch: Semi-supervised learning with contrastive graph
  regularization.
\newblock In {\em Proceedings of the IEEE/CVF International Conference on
  Computer Vision}, pages 9475--9484, 2021.

\bibitem{nlp-prompt-survey}
Pengfei Liu, Weizhe Yuan, Jinlan Fu, Zhengbao Jiang, Hiroaki Hayashi, and
  Graham Neubig.
\newblock Pre-train, prompt, and predict: A systematic survey of prompting
  methods in natural language processing.
\newblock {\em arXiv preprint arXiv:2107.13586}, 2021.

\bibitem{sift}
David~G Lowe.
\newblock Object recognition from local scale-invariant features.
\newblock In {\em Proceedings of the seventh IEEE international conference on
  computer vision}, volume~2, pages 1150--1157. Ieee, 1999.

\bibitem{aircraft}
Subhransu Maji, Esa Rahtu, Juho Kannala, Matthew Blaschko, and Andrea Vedaldi.
\newblock Fine-grained visual classification of aircraft.
\newblock {\em arXiv preprint arXiv:1306.5151}, 2013.

\bibitem{semi-eval}
Avital Oliver, Augustus Odena, Colin~A Raffel, Ekin~Dogus Cubuk, and Ian
  Goodfellow.
\newblock Realistic evaluation of deep semi-supervised learning algorithms.
\newblock {\em Advances in neural information processing systems}, 31, 2018.

\bibitem{consensus-knn}
Vittal Premachandran and Ramakrishna Kakarala.
\newblock Consensus of k-nns for robust neighborhood selection on graph-based
  manifolds.
\newblock In {\em Proceedings of the IEEE conference on computer vision and
  pattern recognition}, pages 1594--1601, 2013.

\bibitem{silhouette}
Peter~J Rousseeuw.
\newblock Silhouettes: a graphical aid to the interpretation and validation of
  cluster analysis.
\newblock {\em Journal of computational and applied mathematics}, 20:53--65,
  1987.

\bibitem{finch}
Saquib Sarfraz, Vivek Sharma, and Rainer Stiefelhagen.
\newblock Efficient parameter-free clustering using first neighbor relations.
\newblock In {\em Proceedings of the IEEE/CVF Conference on Computer Vision and
  Pattern Recognition}, pages 8934--8943, 2019.

\bibitem{prompt-ensemble}
Timo Schick and Hinrich Sch{\"u}tze.
\newblock Exploiting cloze questions for few shot text classification and
  natural language inference.
\newblock {\em arXiv preprint arXiv:2001.07676}, 2020.

\bibitem{vit}
Gilad Sharir, Asaf Noy, and Lihi Zelnik-Manor.
\newblock An image is worth 16x16 words, what is a video worth?
\newblock {\em arXiv preprint arXiv:2103.13915}, 2021.

\bibitem{herb}
Kiat~Chuan Tan, Yulong Liu, Barbara Ambrose, Melissa Tulig, and Serge Belongie.
\newblock The herbarium challenge 2019 dataset.
\newblock {\em arXiv preprint arXiv:1906.05372}, 2019.

\bibitem{mt}
Antti Tarvainen and Harri Valpola.
\newblock Mean teachers are better role models: Weight-averaged consistency
  targets improve semi-supervised deep learning results.
\newblock {\em Advances in neural information processing systems}, 30, 2017.

\bibitem{tsne}
Laurens Van~der Maaten and Geoffrey Hinton.
\newblock Visualizing data using t-sne.
\newblock {\em Journal of machine learning research}, 9(11), 2008.

\bibitem{ssl-survey}
Jesper~E Van~Engelen and Holger~H Hoos.
\newblock A survey on semi-supervised learning.
\newblock {\em Machine Learning}, 109(2):373--440, 2020.

\bibitem{gcd}
Sagar Vaze, Kai Han, Andrea Vedaldi, and Andrew Zisserman.
\newblock Generalized category discovery.
\newblock In {\em Proceedings of the IEEE/CVF Conference on Computer Vision and
  Pattern Recognition}, pages 7492--7501, 2022.

\bibitem{dataset-cub}
Catherine Wah, Steve Branson, Peter Welinder, Pietro Perona, and Serge
  Belongie.
\newblock The caltech-ucsd birds-200-2011 dataset.
\newblock 2011.

\bibitem{filter-bank}
Yaming Wang, Vlad~I Morariu, and Larry~S Davis.
\newblock Learning a discriminative filter bank within a cnn for fine-grained
  recognition.
\newblock In {\em Proceedings of the IEEE conference on computer vision and
  pattern recognition}, pages 4148--4157, 2018.

\bibitem{hungarian}
MB Wright.
\newblock Speeding up the hungarian algorithm.
\newblock {\em Computers \& Operations Research}, 17(1):95--96, 1990.

\bibitem{dec}
Junyuan Xie, Ross Girshick, and Ali Farhadi.
\newblock Unsupervised deep embedding for clustering analysis.
\newblock In {\em International conference on machine learning}, pages
  478--487. PMLR, 2016.

\bibitem{noisy-student}
Qizhe Xie, Minh-Thang Luong, Eduard Hovy, and Quoc~V Le.
\newblock Self-training with noisy student improves imagenet classification.
\newblock In {\em Proceedings of the IEEE/CVF conference on computer vision and
  pattern recognition}, pages 10687--10698, 2020.

\bibitem{ood-survey}
Jingkang Yang, Kaiyang Zhou, Yixuan Li, and Ziwei Liu.
\newblock Generalized out-of-distribution detection: A survey.
\newblock {\em arXiv preprint arXiv:2110.11334}, 2021.

\bibitem{tpg}
Xingwei Yang, Lakshman Prasad, and Longin~Jan Latecki.
\newblock Affinity learning with diffusion on tensor product graph.
\newblock {\em IEEE transactions on pattern analysis and machine intelligence},
  35(1):28--38, 2012.

\bibitem{s4l}
Xiaohua Zhai, Avital Oliver, Alexander Kolesnikov, and Lucas Beyer.
\newblock S4l: Self-supervised semi-supervised learning.
\newblock In {\em Proceedings of the IEEE/CVF International Conference on
  Computer Vision}, pages 1476--1485, 2019.

\bibitem{mixup}
Hongyi Zhang, Moustapha Cisse, Yann~N Dauphin, and David Lopez-Paz.
\newblock mixup: Beyond empirical risk minimization.
\newblock {\em arXiv preprint arXiv:1710.09412}, 2017.

\bibitem{rs++}
Bingchen Zhao and Kai Han.
\newblock Novel visual category discovery with dual ranking statistics and
  mutual knowledge distillation.
\newblock {\em Advances in Neural Information Processing Systems},
  34:22982--22994, 2021.

\bibitem{wscl}
Mingkai Zheng, Fei Wang, Shan You, Chen Qian, Changshui Zhang, Xiaogang Wang,
  and Chang Xu.
\newblock Weakly supervised contrastive learning.
\newblock In {\em Proceedings of the IEEE/CVF International Conference on
  Computer Vision}, pages 10042--10051, 2021.

\bibitem{ncl}
Zhun Zhong, Enrico Fini, Subhankar Roy, Zhiming Luo, Elisa Ricci, and Nicu
  Sebe.
\newblock Neighborhood contrastive learning for novel class discovery.
\newblock In {\em Proceedings of the IEEE/CVF Conference on Computer Vision and
  Pattern Recognition}, pages 10867--10875, 2021.

\bibitem{openmix}
Zhun Zhong, Linchao Zhu, Zhiming Luo, Shaozi Li, Yi Yang, and Nicu Sebe.
\newblock Openmix: Reviving known knowledge for discovering novel visual
  categories in an open world.
\newblock In {\em Proceedings of the IEEE/CVF Conference on Computer Vision and
  Pattern Recognition}, pages 9462--9470, 2021.

\bibitem{ibot}
Jinghao Zhou, Chen Wei, Huiyu Wang, Wei Shen, Cihang Xie, Alan Yuille, and Tao
  Kong.
\newblock ibot: Image bert pre-training with online tokenizer.
\newblock {\em arXiv preprint arXiv:2111.07832}, 2021.

\bibitem{local-agg}
Chengxu Zhuang, Alex~Lin Zhai, and Daniel Yamins.
\newblock Local aggregation for unsupervised learning of visual embeddings.
\newblock In {\em Proceedings of the IEEE/CVF International Conference on
  Computer Vision}, pages 6002--6012, 2019.

\end{thebibliography}
}

\clearpage

\appendix

\twocolumn[
\begin{@twocolumnfalse}
\section*{\centering {\Large PromptCAL: Contrastive Affinity Learning via Auxiliary Prompts for Generalized Novel Category Discovery -- {\em Supplementary Material}}}
\end{@twocolumnfalse}
]
\vspace{30pt}
    
\section*{Appendix}
In this appendix, we further provide detailed descriptions on the following contents:
\begin{itemize}
\setlength{\itemsep}{0pt}
    \item Additional details on our SemiAG method in Appendix~\ref{app:semi-ag}.
    \item Dataset profiles in Appendix~\ref{app:data}.
    \item The complete implementation details in Appendix~\ref{app:implement}.
    \item Additional experimental results in Appendix~\ref{app:exp}.
    \item Training algorithm of PromptCAL in Appendix~\ref{app:algorithm}.
    \item Qualitative and visualization results in Appendix~\ref{app:vis}.
    \item Efficiency analysis in Appendix~\ref{app:efficiency}.
    \item Broader impact and limitations in Appendix~\ref{app:limitations}.
    \item License for experimental datasets in Appendix~\ref{app:lic}.
\end{itemize}

\section{Additional details on SemiAG}
\label{app:semi-ag}
In this section, we present an extended description of TPG~\cite{tpg} affinity propagation algorithm that underlies our SemiAG method. \par 
Suppose we have a graph $G=(V, \mathbf{E})$ with a node set $V$ and an edge set $\mathbf{E}$. 
In our context, $V$ is a set of $N$ embeddings and $\mathbf{E} \in \mathbf{R}^{N \times N}$ represents the pairwise affinity matrix.
TPG runs a graph diffusion process on a tensor product graph $\mathcal{G}=(V \times V, \mathcal{E})$ defined on $G$, where $\mathcal{E}=\mathbf{E} \otimes \mathbf{E}$ represents a $4$-dim tensor. In particular, for $i,j,k,l=1...,N$, the tensor element $\mathcal{E}_{i,j,k,l}=\mathbf{E}_{i,j}\mathbf{E}_{k,l} \in \mathbf{R}^{NN \times NN}$. In other words, the tensor graph $\mathcal{G}$ can be intuitively considered as a higher-order graph through cartesian product between $G$ and itself. Then the graph diffusion process on $\mathcal{G}$ is formulated as:
\begin{equation}
    \mathcal{E}^{(t)} = \sum_{i=0}^{t}{\mathcal{E}^i} \notag
\end{equation}
where $\mathcal{E}^{(t)}$ denotes the $t$-th step affinity matrix and $\mathcal{E}^i$ is \mbox{$i$-power} of $\mathcal{E}$. Theoretically, if the row-sum of $\mathcal{E}$ is less than one, $\mathcal{E}^{(t)}$ will converge to a nontrivial solution. 
To make computation tractable on large-scale data, TPG~\cite{tpg} proposes an iterative equation without multiplication on tensors which theoretically guarantees the same converged solution, which is formulated as:
\begin{equation}
    \mathbf{Q}^{(t+1)} = \mathbf{E} \mathbf{Q}^{(t)} \mathbf{E}^T + \mathbf{I} \notag
\end{equation}
where $\mathbf{I}$ denotes an identity matrix, $\mathbf{E}$ is the affinity matrix, and $\mathbf{Q}^{(0)}=\mathbf{E}$. \par 

In our work, we calibrates the affinity graph with only first-order structural information and, thus, set the diffusion step $\eta=1$ since: firstly, online diffusion till convergence at each iteration will incur great computation overheads; besides, we find larger diffusion steps will include noisy false positives which significantly degrades the overall performance. 
Based on our further observation that the row-wise sum constraint has negligible effect on final performance, we exclude the row-wise sum threshold in TPG~\cite{tpg} as another hyperparameter. \par 

\section{Dataset details}
\label{app:data}
We evaluate \mbox{PromptCAL} on six benchmarks, \ie, CIFAR-10~\cite{cifar}, CIFAR-100~\cite{cifar}, ImageNet-100~\cite{imagenet}, CUB-200~\cite{dataset-cub}, StandfordCars~\cite{scars}, and Aircraft~\cite{aircraft}.
The profile of six benchmark datasets is displayed in Table \ref{tab:dataset}. Our dataset splits follow GCD~\cite{gcd}. \par

\begin{table}[H]
    \centering
    \resizebox{\linewidth}{!}{
    \begin{tabular}{l|cccccc}
    \toprule
        \textbf{Dataset} & CIFAR-10 & CIFAR-100 & ImageNet-100 & CUB-200 & Aircraft & StanfordCars \\
    \midrule
        \#Images in $\mathcal{D}$ & 50k & 50k & 127.2k & 6k & 6.6k & 8.1k \\
        \#Classes ($|\mathcal{C}|$) & 10 & 100 & 100 & 200 & 100 & 196 \\
        \#Known Classes ($|\mathcal{C}_{kwn}|$) & 5 & 80 & 50 & 100 & 50 & 98 \\
    \bottomrule
    \end{tabular}
    }
    \caption{\textbf{The dataset profiles of six benchmarks for evaluation.}}
    \label{tab:dataset}
\end{table}

\section{Implementation details}
\label{app:implement}
\noindent \textbf{Architecture and optimization.} Following \cite{gcd}, we use a 12-layer base vision transformer~\cite{vit} with a patch size of 16 (ViT-B/16) as our backbone in all experiments. The backbone weights are initialized with pre-trained DINO~\cite{DINO} on the ImageNet-1K~\cite{imagenet} dataset. The first $11$ blocks of the backbone are frozen as in \cite{gcd}.
For our PromptCAL, we further adapt pre-trained ViT \cite{vit} with prompts by prepending $5$ prompts before each block (in VPT-Deep scheme~\cite{vpt}). We only supervise the first $2$ of $5$ prompts at the last block with DPR loss, and all remaining prompts are unsupervised and thus automatically learned.
In practice, this ViT backbone can be of any architecture and pre-trained with any self-supervised learning method on large-scale datasets.
Initially, we separately adopt two DINO~\cite{DINO} projection heads for {\tt [CLS]} and {\tt [P]} to avoid negative interferences, which are randomly initialized.
In both stages, we fix the batch size to $128$ on all datasets; besides, we optimize PromptCAL with standard SGD with a momentum of $0.9$, a weight decay of $5 \times 10^{-5}$, and an initial learning rate of $0.1$.
For all datasets, we train PromptCAL with $200$ epochs in the first stage; in the second stage, we train PromptCAL with $70$ epochs on CIFAR-10/100 and ImageNet-100 datasetes; while, we optimize PromptCAL by $100$ epochs on CUB-200, StanfordCars, and Aircraft datasets. \par 

\noindent \textbf{Warmup training.}
In the $1^{st}$ stage training of \mbox{PromptCAL}, we adopt an unsupervised $L_2$ distillation loss on ImageNet-1K~\cite{imagenet} with a loss weight of $\min \big(0,0.5 \times (1-\frac{E}{5})\big)$. Here, $E$ denotes the epoch number.
We add this loss based on consideration of potential adverse effects of randomly initialized visual prompts on the class token. \par 

\noindent \textbf{Contrastive affinity learning.}
In the $2^{nd}$ stage training of \mbox{PromptCAL}, model parameters (prompt-adapted backbone with two heads) are initialized by the best warmed-up checkpoint at the $1^{st}$ stage.
For SemiAG parameters, we fix the neighborhood size $K=|\mathcal{M}|/(4|\mathcal{C}|)$ for all datasets unless otherwise specified.
We fix sizes of both memories as $|\mathcal{M}|=|\mathcal{M}_{\text{P}}|=4096$ and set $N_{neg}=1024$ in all experiments.
Furthermore, since most edges of the binarized affinity graph $\mathbf{G}_b^\prime$ are of small values, we first compute the mean value of non-zero affinities; then, we fix threshold $q$ to $80\%$ quantile of affinities above this value for all fine-grained datasets, and $50\%$ for all generic datasets.
We fix diffusion step $\eta=1$.
For loss parameters, we fix $\alpha=0.35$, $\tau=1.0$, and $\tau_a=0.07$ based on existing literature~\cite{gcd, moco, DINO}. Besides, we determine $\gamma=0.35$ and $\beta=0.6$ via first and second stage validation scores on the held-out validation set.
Our teacher model is initialized by the student weights at the beginning, and we conduct momentum updates with a momentum of $0.999$ at each iteration.
During the inference, the {\tt [CLS]} representation of the student model is used for prediction. \par 

\noindent \textbf{Validation scheme.}
Follow GCD~\cite{gcd} setup, we assume access to a small validation set, in which only samples from known classes are labeled.
In the first stage, we keep the best checkpoint with the highest clustering accuracy on {\tt Known} on the validation set. In the second stage, we keep the best checkpoint with the highest clustering quality on the validation set for evaluation. We define clustering quality as the average score of the clustering accuracy on {\tt Known} classes and unsupervised Silhouette score~\cite{silhouette} on {\tt New}. Note that there is no information leakage, since Silhouette score does not need ground-truth label. \par

\noindent \textbf{Other baselines.} For GCD~\cite{gcd}, since our dataset splits are consistent with theirs, we report their official scores for main comparisons. In our ablations, we reproduce its results based on their official codes. For ORCA~\cite{orca}, we adapt their backbone from ResNet to the same pre-trained DINO and obtain results based on their official codes. 
For our baseline (PromptCAL w/o prompt), we remove all the prompts and DPR loss on them; besides, we keep the warmup training stage for fair comparison. Other parameters follow the standard setups. \par

\begin{table*}[]
    \centering
    \resizebox{\linewidth}{!}{
    \begin{tabular}{ccccc}
        \includegraphics[scale=0.2]{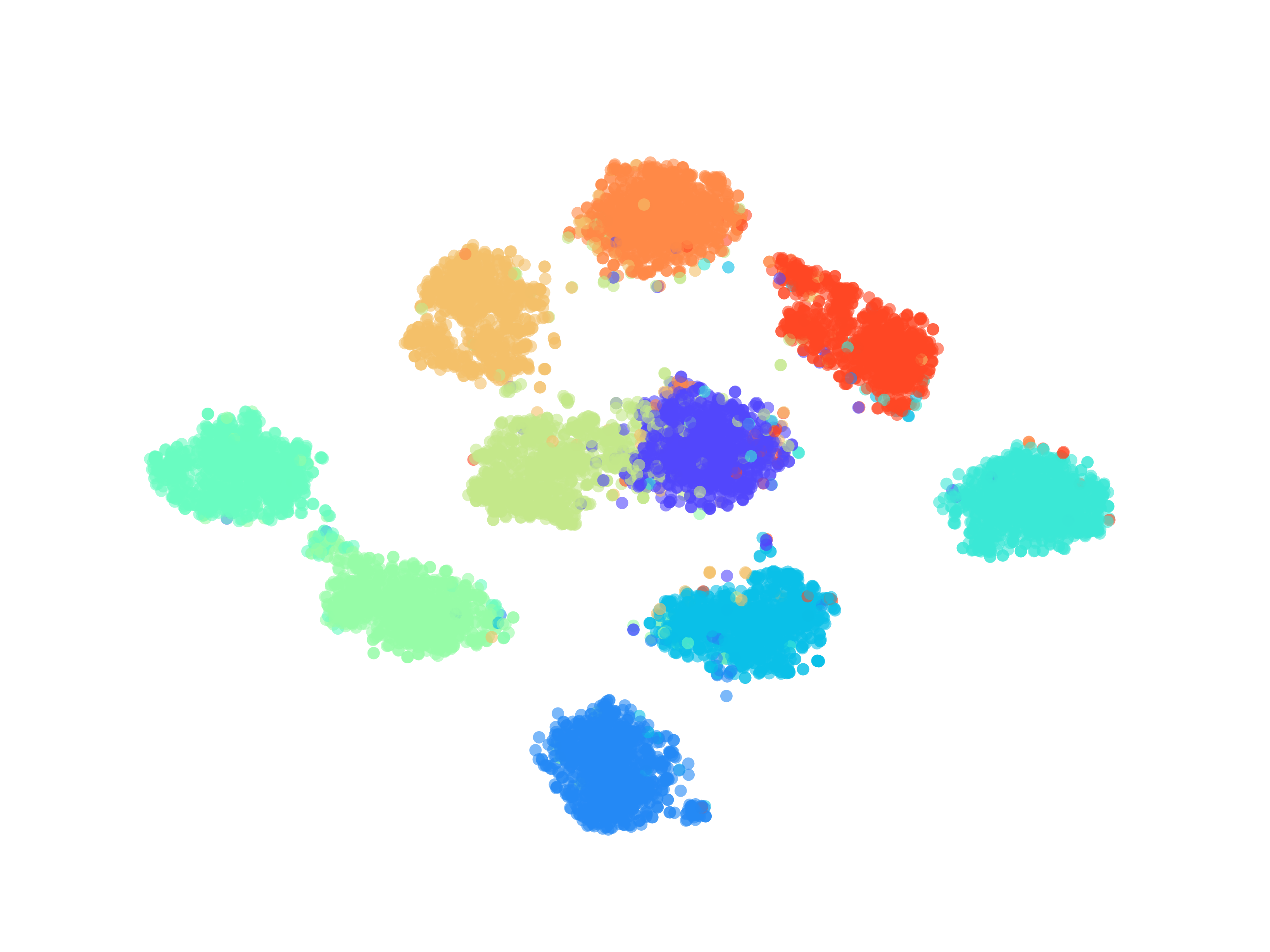} & 
        \includegraphics[scale=0.2]{cifar10-vpt_5-0-test_embedding_all.png} &
        \includegraphics[scale=0.2]{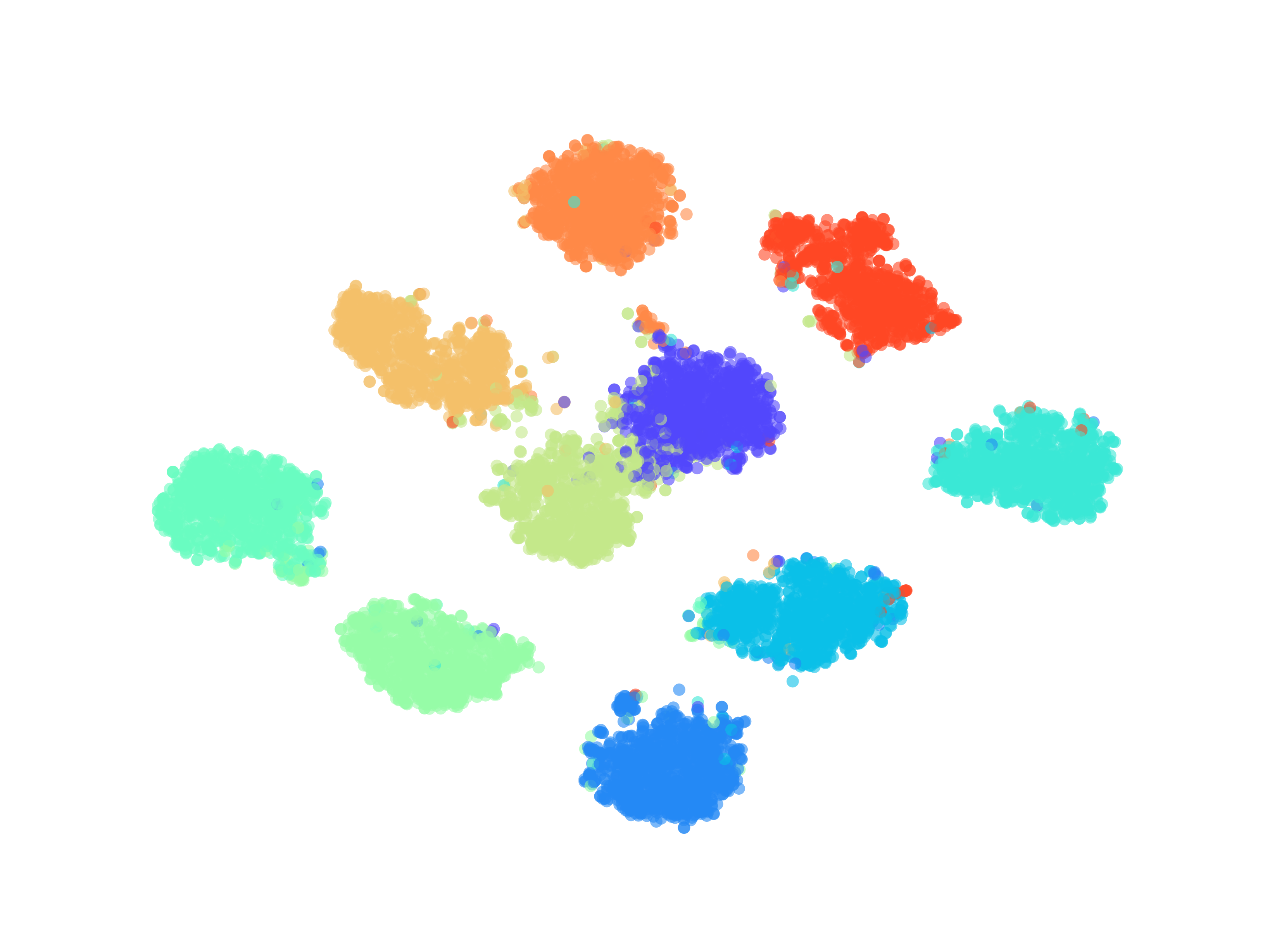} & \includegraphics[scale=0.2]{cifar10-PromptCAL_2-0-test_embedding_all.png} & \includegraphics[scale=0.08]{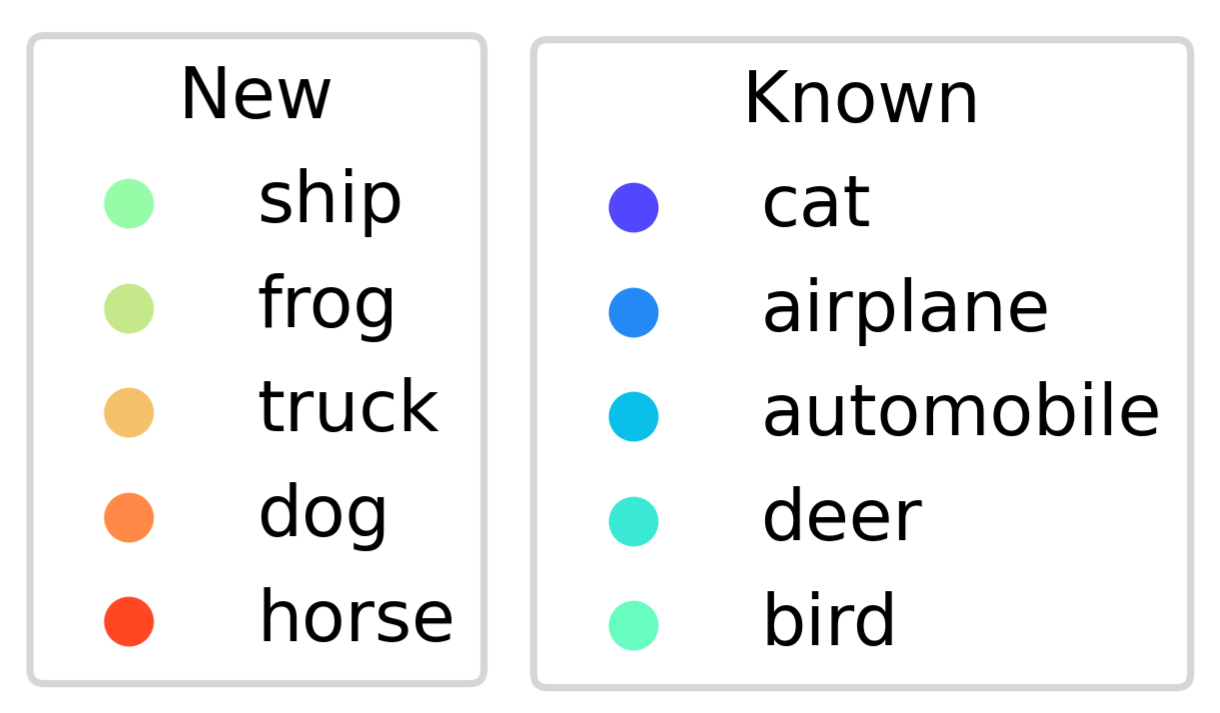} \\
        (a) GCD {\tt [CLS]} & (b) VPT-5 {\tt [CLS]} & (c) PromptCAL-$1^{st}$ {\tt [CLS]} & (d) PromptCAL-$2^{nd}$ {\tt [CLS]} & (e) \\
        \includegraphics[scale=0.2]{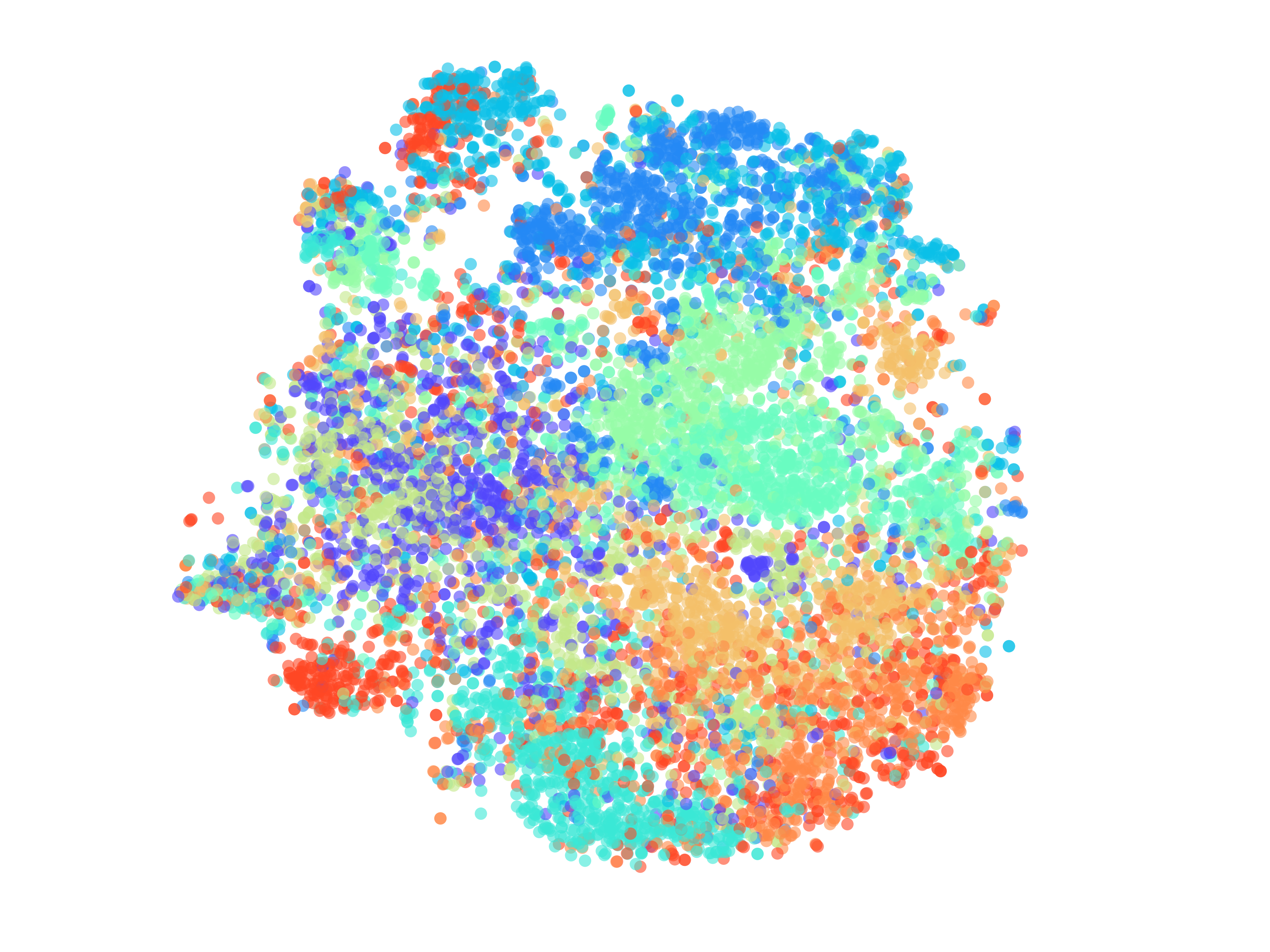} & 
        \includegraphics[scale=0.2]{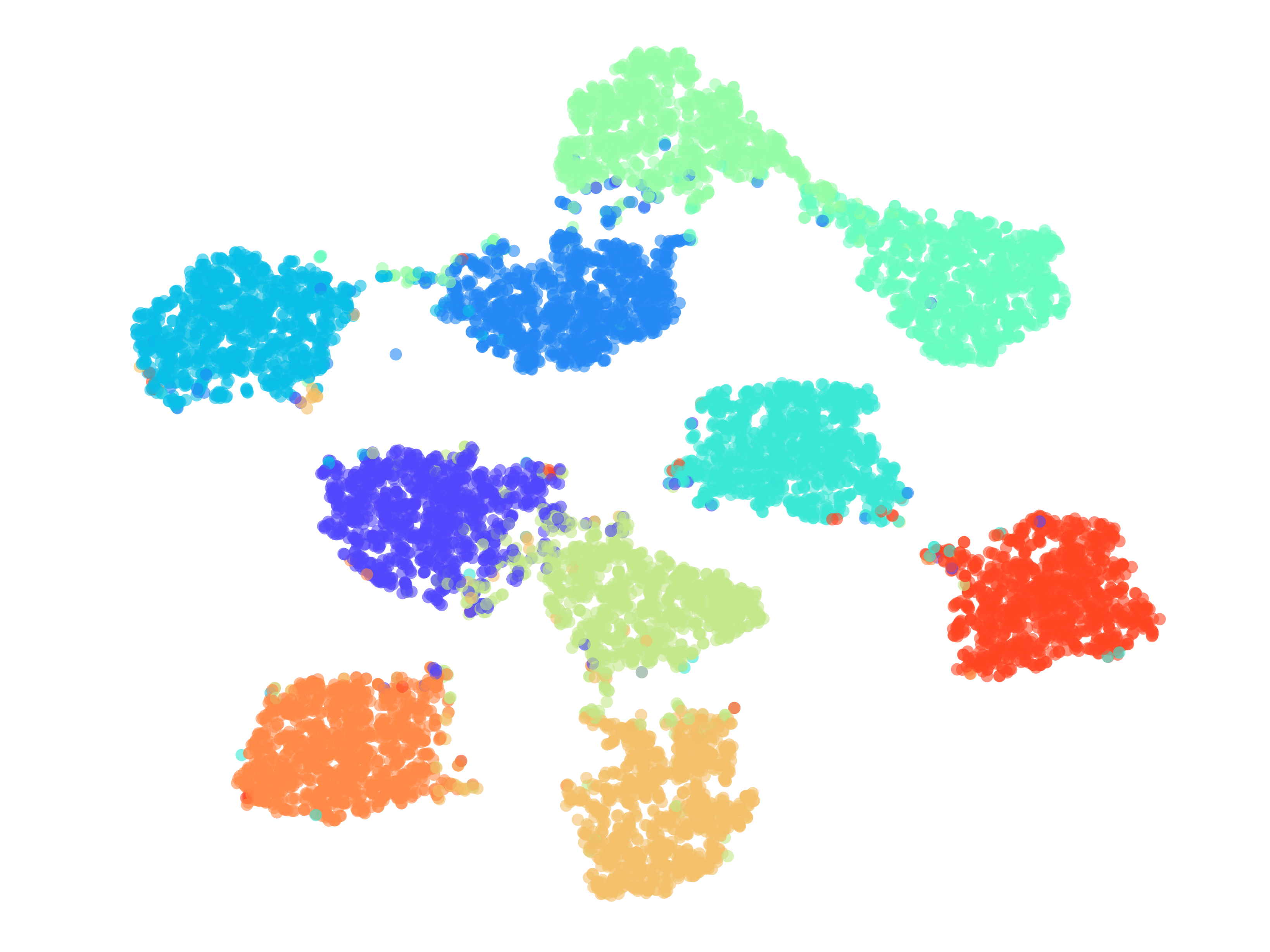} & 
        \includegraphics[scale=0.2]{cifar10-PromptCAL_2-1-test_embedding_all.png} &
        \includegraphics[scale=0.2]{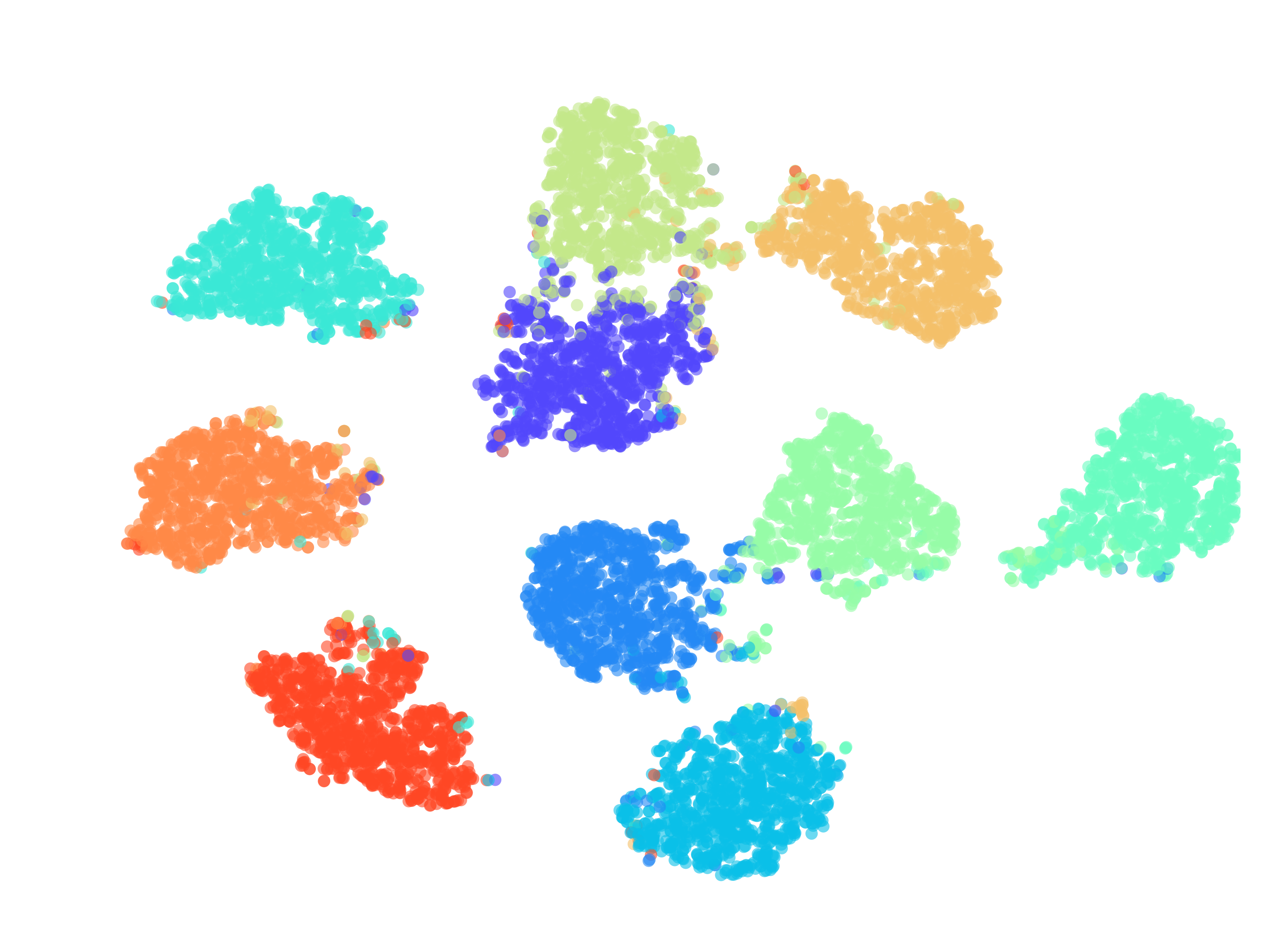} &
        \includegraphics[scale=0.2]{cifar10-PromptCAL_2-3-test_embedding_all.png} \\
        (f) VPT-5 {\tt [P]}$^*$ & (g) PromptCAL-$1^{st}$ {\tt [P]} & (h) PromptCAL-$2^{nd}$ {\tt [P]} & (i) PromptCAL-$1^{st}$ first {\tt [P]}$^*$ & (j) PromptCAL-$2^{nd}$ second {\tt [P]}$^*$ \\
    \end{tabular}
    }
    \caption{\textbf{The t-SNE~\cite{tsne} visualization of ViT embeddings on CIFAR-10 test set} for GCD~\cite{gcd}, naive VPT model~\cite{vpt}, and PromptCAL-$1^{st}$ stage and $2^{nd}$ stage,
    Here, {\tt [CLS]}, {\tt [P]}, and {\tt [P]}$^*$ denote embeddings from ViT class token, ensembled prompts supervised by DPR loss, and unsupervised prompts, respectively.
    The embedding clustering shows that DPR reinforces the semantic discriminativeness of {\tt [P]}, and for {\tt [P]}$^*$ despite no explicit supervision. 
    (e) exhibits the class name each color denotes.
    All figures share the same axis scale.}
    \label{tab:mpc-vis}
\end{table*}

\section{Additional experiment results}
\label{app:exp}


\subsection{Inductive category discovery}
\label{app:compare-test}
In contrast to the evaluation protocol on transductive category discovery GCD~\cite{gcd}, we also conduct ablation experiments on inductive category discovery protocol proposed in ORCA~\cite{orca}. In other words, besides achieving high performance on category discovery on the unlabeled training data (transductive protocol), we also expect models to learn general rules applied to unseen test sets (inductive protocol).
Therefore, we conduct experiment under this inductive evaluation protocol on three benchmarks (CUB-200~\cite{dataset-cub}, CIFAR-100~\cite{cifar}, and ImageNet-100~\cite{imagenet} datasets). In this experiment, we hold out $10\%$ (labeled and unlabeled) training data as the validation set for GCD and PromptCAL.
From displayed results in Table~\ref{tab:generalize}, we can conclude that our PromptCAL achieves the best performance on three datasets, which manifests its good generalization capability. Meanwhile, we observe that PromptCAL boosts performance on {\tt New} with significant margins. \par 

\subsection{Additional ablation on SemiAG and DPR}
\label{app:additional-ablation}
To further validate the effectiveness of our SemiAG, we conduct ablation on different positive mining methods integrated into our online contrastive learning framework with CAL. 
Besides, we supplement more ablation results on larger datasets (\ie, CIFAR-100 and ImageNet-100 datasets) to showcase that learning with semantically discriminative prompts can achieve notable improvements across various datasets. 
The experiment results are presented in Table~\ref{tab:promp-ablation+}.
Firstly, we notice that SemiAG significantly outperforms other positive mining methods, \ie, naive KNN with SemiPriori (KNN w/ S.P.) and Ranking Statistics (R.S.) \cite{rs}.
The results unveil that both KNN with SemiPriori and RankingStats fail to reliably uncover the substantial semantic information in embedding spaces, which proves that our SemiAG method is the most effective in this open-set setting.
On the other hand, removing either DPR loss or entire prompt-related components in PromptCAL causes noticeable performance drop, \eg, nearly $3\%$ and $2\%$ drops on {\tt All} on CIFAR-100 dataset after removing prompts and DPR loss. Moreover, removing either component also leads to severe overfitting on {\tt Known} classes. \par

\subsection{Visualization on embeddings}
\label{app:vis-mpc}
To inspect the learned semantic discriminativeness of PromptCAL, we visualize embeddings by t-SNE~\cite{tsne} algorithm in Fig.~\ref{tab:mpc-vis}.
Firstly, by comparing (a-d), we can conclude that \mbox{PromptCAL} can effectively learn better semantic clustering, witnessed by higher purity, larger inter-class separation, and high compactness.
Notice in (b) that naive VPT model suffer from degraded clustering performance compared with (a) baseline, which again proves that lack of semantic supervision is a critical issue (see ablations in main content) in prompt tuning.
Interestingly, though not supervised, automatically learned prompts {\tt [P]}$^*$ in (i) and (j) can still learn robust semantically meaningful representation, benefiting from DPR on {\tt [P]}. Meanwhile, DPR loss reinforce this effect in (g) and (h).
Furthermore, we also observe that {\tt [P]} supervised by CAL loss (h) can learn better semantic clustering than those supervised by SemiCL (g), and better benefit {\tt [P]}$^*$ (j).
Thanks to better semantic information supplied by CAL loss, {\tt [CLS]} of PromptCAL-$2^{nd}$ learns more compact and better-separated clusters compared with that of PromptCAL-$1^{st}$.
To summarize the above, we can conclude that the second stage enhances the prompts potential using CAL loss, which further enables prompts and CAL to synergistically improve the overall performance. \par

\begin{figure*}[]
    \centering
    \includegraphics[scale=0.53]{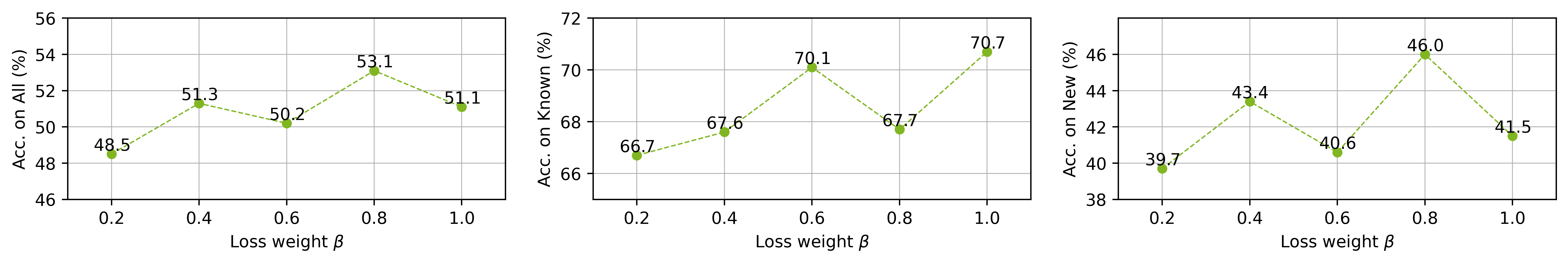}
    \vspace{-10pt}
    \caption{\textbf{Ablation study on the CAL loss weight $\beta$ on StanfordCars~\cite{scars} dataset.}}
    \label{fig:beta-2}
\end{figure*}

\begin{table}[]
    \centering
    \begin{tabular}{c||c|ccc}
    \toprule
        \textbf{Dataset} & \textbf{Setup} & \textbf{All} & \textbf{Known} & \textbf{New} \\
        \midrule
        CUB-200 & w/o prompt & 60.3 & 64.8 & 58.0 \\
        CUB-200 & w/o DPR & 59.3 & 63.3 & 57.4 \\
        CUB-200 & KNN w/ S.P. & 60.1 & \textbf{70.1} & 55.1 \\
        CUB-200 & R.S. & 55.6 & 66.0 & 50.3 \\
        \rowcolor{Green}
        CUB-200 & \mbox{PromptCAL} & \textbf{62.9} & 64.4 & \textbf{62.1} \\
        \midrule
        CIFAR-100 & w/o prompt & 78.1 & 83.0 & 68.4 \\
        CIFAR-100 & w/o DPR & 79.0 & 83.4 & 70.3 \\
        CIFAR-100 & KNN w/ S.P. & 78.7 & 85.3 & 65.4 \\
        CIFAR-100 & R.S. & 75.9 & \textbf{87.1} & 53.4 \\
        \rowcolor{Green}
        CIFAR-100 & \mbox{PromptCAL} & \textbf{81.2} & 84.2 & \textbf{75.3} \\
        \midrule
        ImageNet-100 & w/o prompt & 81.8 & 94.7 & 75.3 \\
        ImageNet-100 & w/o DPR & 80.7 & 94.8 & 73.6 \\
        ImageNet-100 & KNN w/ S.P. & 81.9 & 95.0 & 75.3 \\
        ImageNet-100 & R.S. & 78.1 & \textbf{95.2} & 69.4 \\
        \rowcolor{Green}
        ImageNet-100 & \mbox{PromptCAL} & \textbf{83.1} & 92.7 & \textbf{78.3} \\
        \bottomrule
    \end{tabular}
    \caption{\textbf{Further ablation study on CUB-200~\cite{dataset-cub}, CIFAR-100~\cite{cifar}, and ImageNet-100~\cite{imagenet} datasets.} We investigate four setups: the first is PromptCAL removing all prompt related components; the second is PromptCAL without DPR loss; the third is replacing SemiAG with naive KNN incorporated with SemiPriori; the last one is replacing our SemiAG with RankingStats~\cite{rs} pseudo labeling.}
    \label{tab:promp-ablation+}
\end{table}

\begin{table}[]
    \centering
    \begin{tabular}{c||ccc|ccc}
    \toprule
     & \multicolumn{3}{c|}{CIFAR-100} & \multicolumn{3}{c}{Aircraft} \\
        $\mathbf{K}$ & \textbf{All} & \textbf{Known} & \textbf{New} & \textbf{All} & \textbf{Known} & \textbf{New} \\
    \midrule
        5 & 80.9 & \textbf{85.5} & 71.7 & 49.0 & 54.4 & 46.3 \\
        10 &\textbf{81.2} & 84.2 & 75.3 & \textbf{52.2} & 52.2 & \textbf{52.3} \\
        15 & 80.2 & 83.4 & 74.0 & 50.6 & \textbf{55.1} & 48.4 \\
        20 & 78.9 & 80.3 & \textbf{76.1} & 47.4 & 52.5 & 45.0 \\
    \bottomrule
    \end{tabular}
    \caption{\textbf{Ablation study on the neighborhood size $K$ on the CIFAR-100~\cite{cifar} and Aircraft~\cite{aircraft} datasets.}}
    \label{tab:neighorbood}
\end{table}


\subsection{Sensitivity analysis on hyper-parameters.}
\label{app:exp-sensitivity}
We conduct ablation experiments on critical hyper-parameters of \mbox{PromptCAL}, which includes: {\tt (1)} CAL loss weight $\beta$; {\tt (2)} neighborhood size $K$; {\tt (3)} different pretraining methods; {\tt (4)} number of auxiliary prompts. \par 

\noindent \textbf{CAL loss weight.} We sample $\beta$ values from $0.2$ to $1.0$ at an interval of $0.2$ and run experiments on StanfordCars dataset. The results are visualized in Fig.~\ref{fig:beta-2}. We observe that decreased weights of contrastive affinity learning will cause model suffer from low performance on {\tt New}. 
We argue that, although different datasets exhibit different trends, the model performance is fairly robust within the modest value range (from $0.4$ to $0.8$). \par

\noindent \textbf{Neighborhood size.} We select $K=5,10,15,20$ for ablations on two datasets (CIFAR-100 and Aircraft, both with $100$ {\tt All} classes). Results in Table~\ref{tab:neighorbood} display that PromptCAL is robust to small $K$; while, its performance degrades largely as the neighborhood expands. We guess it is because false positive has severer negative effects than false negatives. \par

\noindent \textbf{Pretraining.} We argue that PromptCAL can take advantage of the property of the high KNN precision of ViT, which are pre-trained in various schemes. In Table~\ref{tab:dino-1}, we replace DINO~\cite{DINO} pre-trained ViT with iBoT~\cite{ibot} pre-trained ViT as our backbone in CIFAR-100 experiments \footnote{The KNN precision of DINO and iBoT on ImageNet-1K dataset are $76.1\%$ and $77.1\%$, respectively \cite{ibot}.}. We can show that PromptCAL further improves as iBoT possesses higher KNN precision~\cite{ibot}. It manifests that our PromptCAL performance is likely to correlate with better initial representations. \par 

\noindent \textbf{Number of supervised prompts.} We varies the number of supervised prompts to observe sensitivity of performance \wrt this parameter. Table~\ref{tab:supervised-prompt} showcases the results under different setups. We can observe that leaving some unsupervised prompt to learn can provide extra flexibility to the backbone and thus achieves the best performance, especially on {\tt New}. In general, PromptCAL is robust to different numbers of supervised prompts. \par

\subsection{Additional results on Herbarium dataset}
We also present evaluation results on the challenging Herbarium2019 \cite{herb} dataset, which consists of 683 classes and 34k images in total. Our dataset split follows \cite{gcd}. Specifically, we set labeling ratio to $50\%$ and known class number to $341$. We compare PromptCAL with other SOTAs on this dataset. Considering larger class numbers, we enlarge the memory size to $2\times 10^{4}$ and $N_{\text{neg}}=5000$, accordingly. We set $K=|\mathcal{M}|/(4|\mathcal{C}|) \approx 7$ in this case. Other parameters follow the setup on fine-grained datasets.
Table~\ref{tab:herb} display the results, which demonstrates our PromptCAL also excels at discovering categories on large vocabulary fine-grained datasets, especially on {\tt New} classes. \par

\begin{table}[]
    \centering
    \begin{tabular}{c|ccc}
    \toprule
        \textbf{Method} & \textbf{All} & \textbf{Known} & \textbf{New} \\
    \midrule
        KMeans~\cite{kmeanspp} & 12.9 & 12.9 & 12.8 \\
        RankStats+~\cite{rs} & 27.9 & \textbf{55.8} & 12.8 \\
        UNO+~\cite{uno} & 28.3 & 53.7 & 14.7 \\
        GCD~\cite{gcd} & 35.4 & 51.0 & 27.0 \\
        ORCA~\cite{orca} & 25.5 & 34.7 & 15.8 \\
        \rowcolor{Green}
        PromptCAL (our) & \textbf{37.0} & 52.0 & \textbf{28.9} \\
    \bottomrule
    \end{tabular}
    \caption{\textbf{Additional experiments on the Herbarium2019~\cite{herb} dataset.}}
    \label{tab:herb}
\end{table}

\begin{table}[]
    \centering
    \begin{tabular}{c|ccc}
    \toprule
        \textbf{Method} & \textbf{All} & \textbf{Known} & \textbf{New} \\
    \midrule
        GCD~\cite{gcd} & 73.0 & 76.2 & 66.5 \\
        \rowcolor{Green}
        PromptCAL (iBoT~\cite{ibot}) & \textbf{83.0} & \textbf{85.0} & \textbf{78.9} \\
        \rowcolor{Green}
        PromptCAL (DINO~\cite{DINO}) & 81.2 & 84.2 & 75.3 \\
    \bottomrule
    \end{tabular}
    \caption{\textbf{Ablation study on pretraining methods on CIFAR-100~\cite{cifar} dataset.}}
    \label{tab:dino-1}
\end{table}

\begin{table*}[]
    \centering
    \resizebox{\linewidth}{!}{
    \begin{tabular}{c||ccccc|ccccc|ccccc}
    \toprule
     & \multicolumn{5}{c|}{CUB-200} & \multicolumn{5}{c|}{CIFAR-100} & \multicolumn{5}{c}{ImageNet-100} \\
        \textbf{Method} & \textbf{All} & \textbf{Known} & \textbf{New} & \textbf{Known$^*$} & \textbf{New$^*$} & \textbf{All} & \textbf{Known} & \textbf{New} & \textbf{Known$^*$} & \textbf{New$^*$} & \textbf{All} & \textbf{Known} & \textbf{New} & \textbf{Known$^*$} & \textbf{New$^*$} \\
    \midrule
        GCD~\cite{gcd} & 57.5 & 64.5 & 50.6 & 69.2 & 57.6 & 70.1 & 76.8 & 43.5 & 78.7 & 58.2 & 79.7 & 92.7 & 66.7 & 92.7 & 66.9 \\
        ORCA (DINO)~\cite{orca} & 40.7 & 61.2 & 20.2 & \textbf{76.3} & 38.3 & 77.7 & 83.6 & 53.9 & 83.6 & 66.6 & 81.3 & \textbf{94.5} & 68.0 & \textbf{94.5} & 71.1 \\
        \rowcolor{Green}
        PromptCAL (our) & \textbf{62.4} & \textbf{68.1} & \textbf{56.8} & 70.1 & \textbf{60.1} & \textbf{81.6} & \textbf{85.3} & \textbf{66.9} & \textbf{86.2} & \textbf{71.3} & \textbf{84.8} & 94.4 & \textbf{75.2} & 94.4 & \textbf{75.3} \\
    \midrule
    \bottomrule
    \end{tabular}
    }
    \caption{\textbf{Evaluation in the inductive GCD setting\cite{orca} on three benchmarks.} The results are reported in accuracy scores on the test set. Here, we also adopt the task-informed evaluation protocol in \cite{uno, orca}, \ie, {\tt Known}$^*$ and {\tt New}$^*$ are evaluated by separate clustering and Hungarian assignment.}
    \label{tab:generalize}
\end{table*}

\begin{table}[]
    \centering
    \resizebox{\linewidth}{!}{
    \begin{tabular}{c||ccc|ccc}
    \toprule
         & \multicolumn{3}{c|}{Stage 1} & \multicolumn{3}{c}{Stage 2} \\
        \textbf{Method} & \textbf{All} & \textbf{Known} & \textbf{New} & \textbf{All} & \textbf{Known} & \textbf{New} \\
    \midrule
        GCD~\cite{gcd} & 51.3 & 56.6 & 48.7 & - & - & - \\
        \rowcolor{Green}
        DPR-2-5 & 51.1 & 55.4 & \textbf{48.9} & \textbf{62.9} & \textbf{64.4} & \textbf{62.1} \\
        \rowcolor{Green}
        DPR-1-5 & \textbf{51.7} & \textbf{57.2} & \textbf{48.9} & 59.9 & 63.0 & 58.4 \\
        \rowcolor{Green}
        DPR-5-5 & 50.9 & 55.6 & 48.6 & 61.0 & 63.6 & 59.8 \\
    \bottomrule
    \end{tabular}
    }
    \caption{\textbf{Ablation study on prompt numbers of our prompt-adapted ViT backbone.} Evaluation conducted on CUB-200~\cite{dataset-cub} dataset.}
    \label{tab:supervised-prompt}
\end{table}

\section{Training algorithm of PromptCAL}
\label{app:algorithm}
Given a training dataset $\mathcal{D}$, we describe our entire training algorithm of \mbox{PromptCAL} in Algo.~\ref{algo}. 
Before \mbox{PromptCAL} training, we adapt the ImageNet pre-trained ViT backbone $f(\cdot | \theta)$ with prompts into $f(\cdot | \theta, \theta_{\text{P}})$, and randomly initialize two identity heads 
$g(\cdot | \theta_{\text{H}})$ and $g_{\text{P}}(\cdot | \theta_{\text{P}, \text{H}})$ for {\tt [CLS]} and {\tt [P]}, respectively. \par

In the $1^{st}$ stage, we sample a batch of images $\mathbf{X}$ with their corresponding labels $\mathbf{Y}$ at each iteration.
Note that ground-truth labels of unlabeled images are masked in $\mathbf{Y}$.
We obtain {\tt [CLS]} and {\tt [P]} projected features ($\mathbf{Z}, \overline{\mathbf{Z}}_{\text{P}}$) by forwarding $\mathbf{X}$ through backbone and two heads.
Next, we compute SemiCL loss (Eq.~2) on the features based on the class labels and label-or-not information in $\mathbf{Y}$.
All tunable parameters ($\theta$, $\theta_{\text{P}}$, $\theta_{\text{H}}$, $\theta_{\text{P}, \text{H}}$) are updated. \par 

Before the $2^{nd}$ stage training, we initialize two empty embedding memory bank $\mathcal{M}, \mathcal{M}_{\text{P}}$ for {\tt [CLS]} and {\tt [P]}, respectively. Besides, we initialize the teacher model with the student weights.
During the training, for each sampled batch ($\mathbf{X}$, $\mathbf{Y}$), we first obtain student embeddings of {\tt [CLS]} and ensembled {\tt [P]} ($\mathbf{H}, \overline{\mathbf{H}}_{\text{P}}$), and corresponding student features ($\mathbf{Z}$, $\overline{\mathbf{Z}}_{\text{P}}$) by forwarding images to the student. Meanwhile, we acquire the teacher embeddings and features ($\mathbf{H}_T$, $\overline{\mathbf{H}}_{\text{P}, T}$, $\mathbf{Z}_T$, $\overline{\mathbf{Z}}_{\text{P}, T}$) from the teacher, correspondingly. \par 

Further, we construct a sub-graph for a token (line 14 for the class token and line 18 for ensembled prompts) based on its teacher embeddings of the current batch and all embeddings in its corresponding memory.
Given the sub-graph, we sequentially perform three operations of SemiAG to obtain the calibrated binarized affinity graph (line 15 and 19).
For each student embedding, we utilize its teacher embedding counterpart as a query on the affinity graph to acquire its pseudo positive set and pseudo anchor set with randomly sampled pseudo negatives (line 16 and 20).
With these pseudo positive and anchor sets, we compute CAL loss on embeddings of each token (line 17 and 21) by Eq.~7. \par

Along with CAL loss, we also compute SemiCL loss on the projected features; here, we utilize student embeddings as queries and teacher embeddings as keys in the contrastive loss (Eq.~8 and Eq.~9). In other words, for each student embedding, we construct its positive and anchor sets with teacher embeddings and then compute the semi-supervised contrastive loss.
Next, we obtain the total loss for the {\tt [CLS]} token by combining its SemiCL and CAL loss functions (Eq.~9). After adding our DPR counterpart loss on ensembled prompts, we finally get the total loss at this stage (Eq.~10). \par 

At each iteration, all tunable parameters of the student are updated.
Lastly, we update two memories with teacher embeddings of their corresponding token and update momentum teacher model with the updated student model.
Note that for inference, we adopt embeddings from the {\tt [CLS]} token of the student model $f(\cdot | \theta, \theta_{\text{P}})$ for final predictions. \par

\begin{figure*}[t]
    \centering
    \includegraphics[scale=0.85]{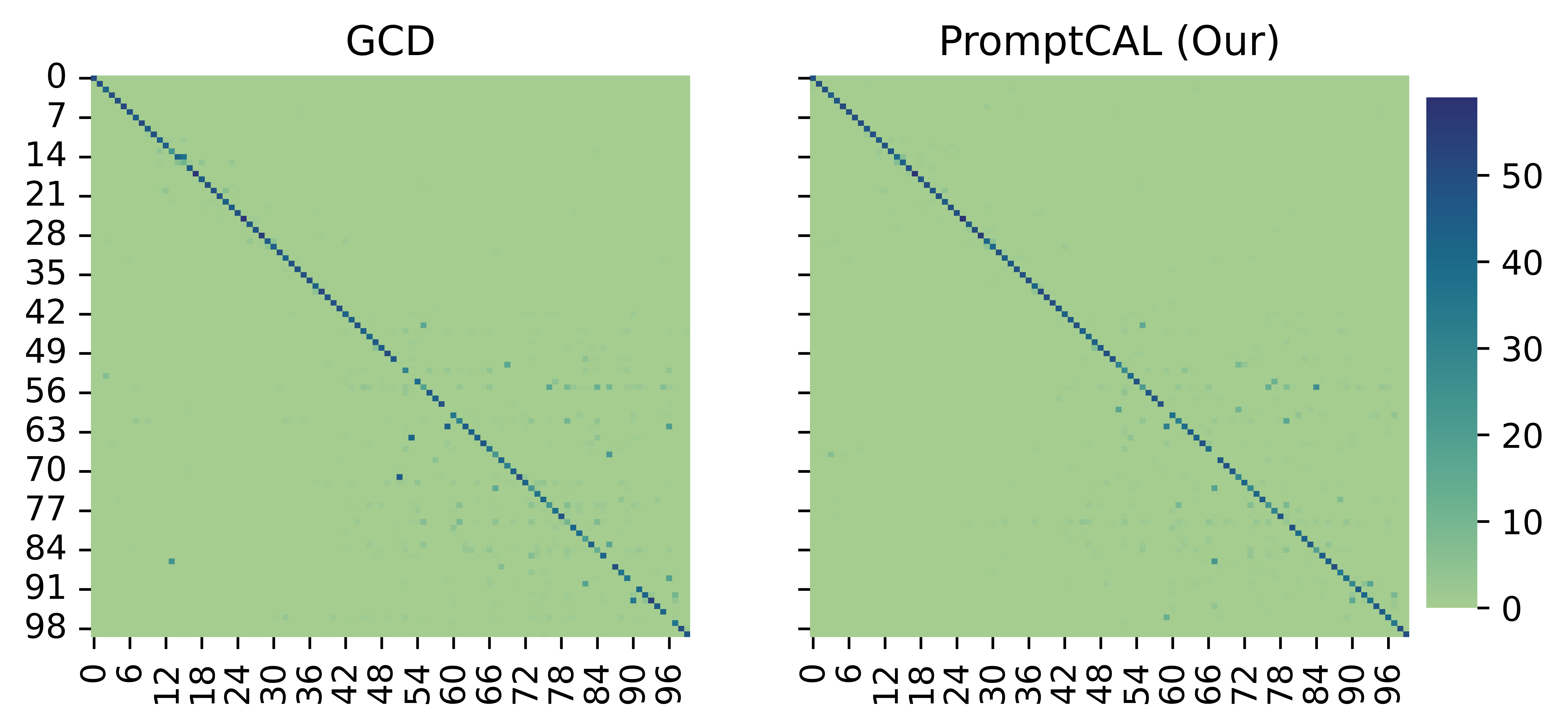}
    \vspace{-10pt}
    \caption{\textbf{Confusion matrix of PromptCAL on ImageNet-100 \cite{imagenet} test set.} The labels on the x-axis and y-axis denotes the class index of our generated split. The first $50$ classes are {\tt Known}, and the last $50$ classes are {\tt New}.}
    \vspace{-6pt}
    \label{fig:confmat}
\end{figure*}

\section{Qualitative results}
\label{app:vis}
In this section, we present qualitative results of categorization confusion matrix, attention map visualization, and KNN retrieval. \par 

\noindent \textbf{Confusion matrix on ImageNet-100.} We present confusion matrix for GCD~\cite{gcd} and our PromptCAL on both {\tt Known} and {\tt New} classes on ImageNet-1K dataset in Fig.~\ref{fig:confmat}. We can observe that our PromptCAL can learn more robust clusters on {\tt New} classes, while preserving high accuracy on {\tt Known}. Moreover, our PromptCAL is less susceptible to confusion between {\tt Known} and {\tt New}. \par

\begin{figure*}[hp]
    \centering
    \includegraphics[scale=0.5]{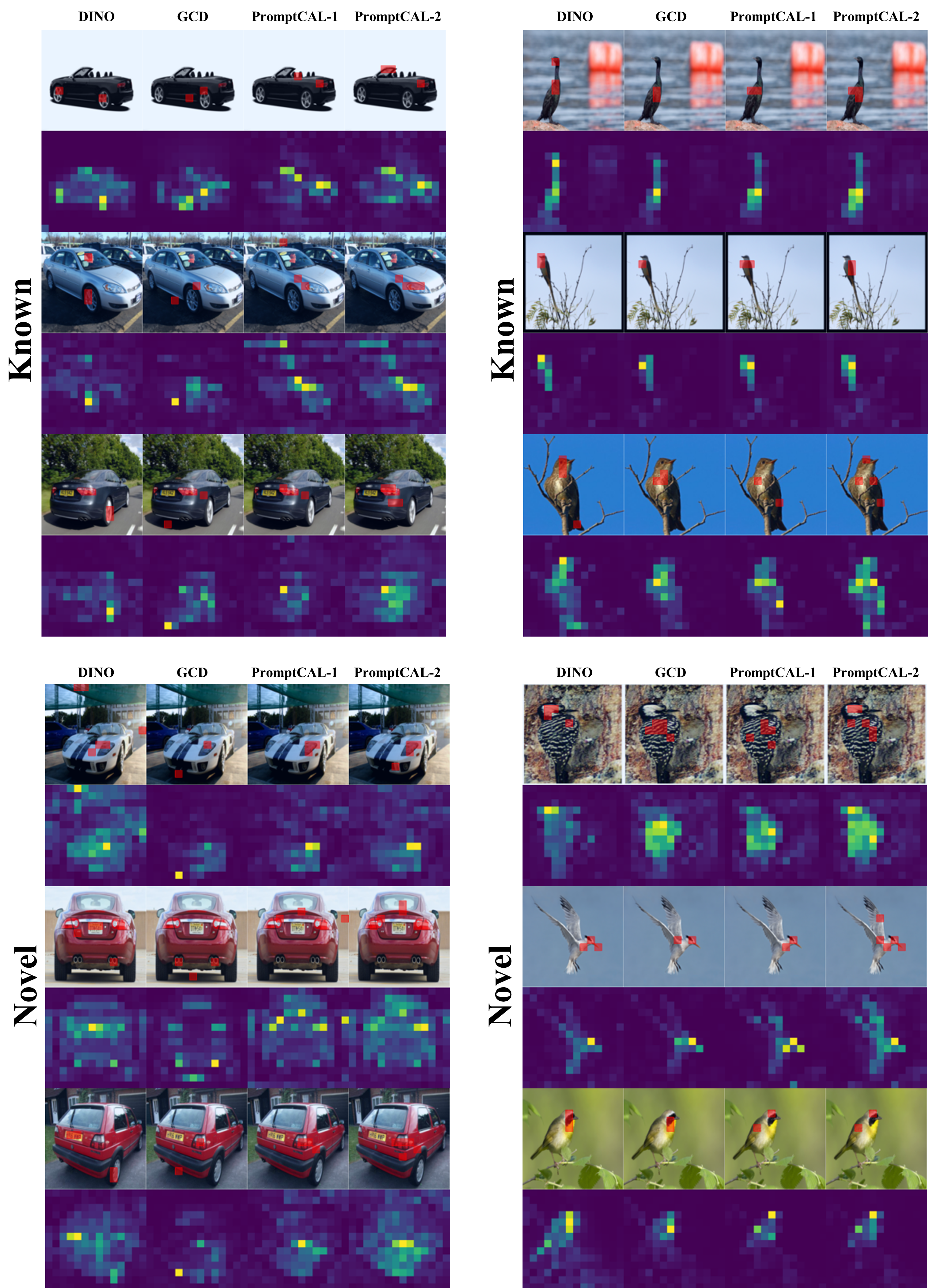}
    \caption{\textbf{Attention map visualization of class tokens for comparison on StandfordCars \cite{scars} (left) and CUB-200 \cite{dataset-cub} (right) datasets.} The columns from left to right refer to attention maps of DINO~\cite{DINO}, GCD~\cite{gcd}, our first stage PromptCAL, and our second stage PromptCAL. In the first row, attended areas are marked in red in each images; the second row display the complete attention maps corresponding to the first row images (yellow regions denote high attention values).}
    \vspace{-5pt}
    \label{tab:attn}
\end{figure*}

\noindent \textbf{Attention map visualization.}
We visualize and compare the attention maps of {\tt [CLS]} tokens of DINO~\cite{DINO}, GCD~\cite{gcd}, PromptCAL-$1^{st}$, and PromptCAL-$2^{nd}$ in Fig.~\ref{tab:attn}. We summarize the following observations: {\tt (1)} DINO attends to the instance discriminative regions, \eg, licence plate, and may overfit on surrounding objects; while, PromptCAL lays more attention on class-specific features, \eg, car lights for cars, and feather textures for birds. {\tt (2)} Although both GCD and PromptCAL can attend to semantically meaningful regions, PromptCAL-$2^{nd}$ focuses on multiple semantically discriminative regions, \eg, car lights and textures, feathers and wings. {\tt (3)} After CAL training, attention maps of PromptCAL-$2^{nd}$ in contrast to that of PromptCAL-$1^{st}$ are remarkably refined. \par

\begin{figure*}[hp]
    \centering
    \includegraphics[scale=0.2]{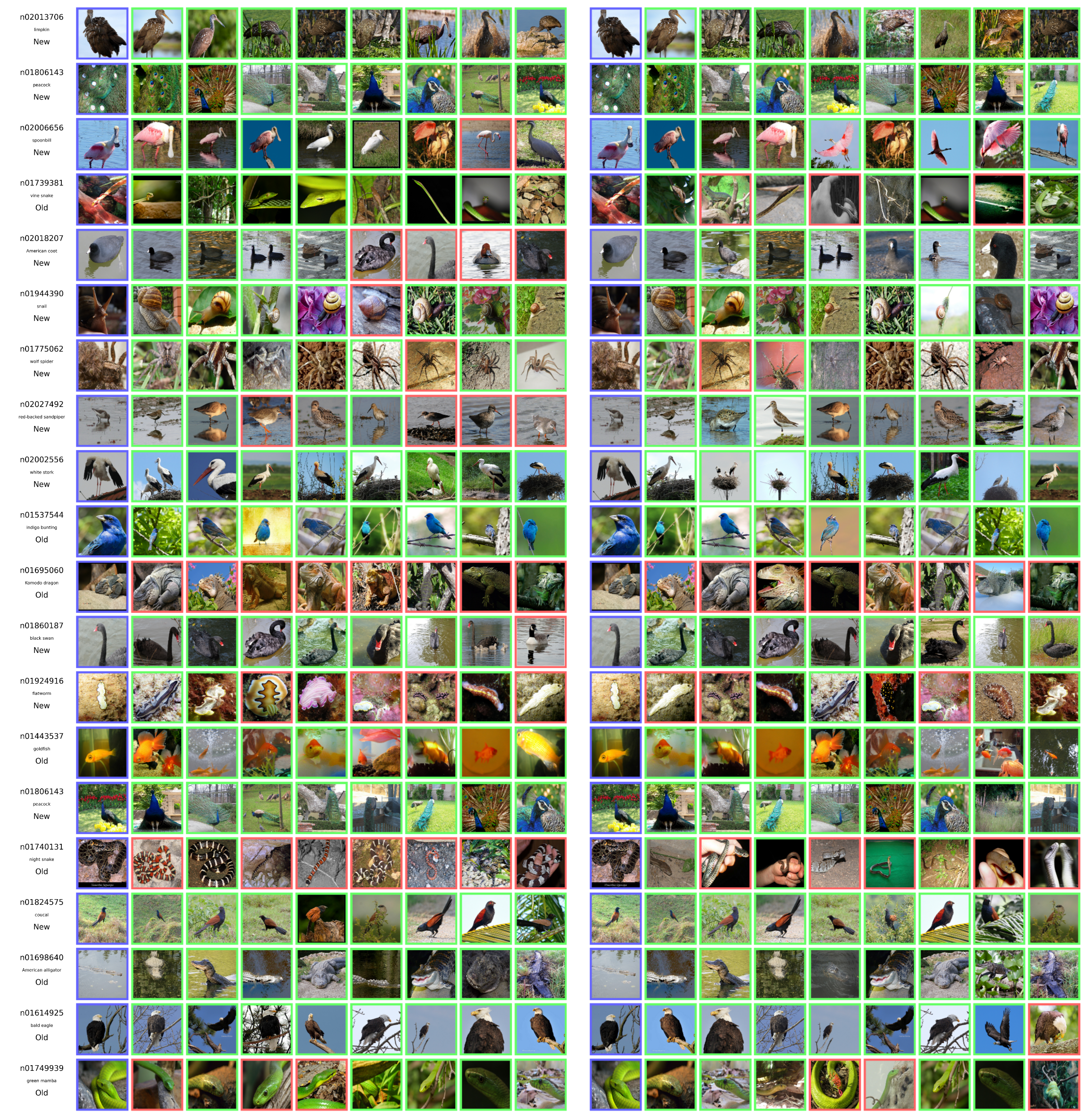}
    \caption{\textbf{Visualization of retrieved $8$-NN for $20$ randomly selected query images (with blue borders).} 
    The correct/incorrect predictions are marked with green/red borders.
    The predictions on the left come from GCD, and the right is from \mbox{PromptCAL}.
    The first column contains ImageNet synsetIDs, category name, and {\tt Known}/{\tt New} for each query. Better view with zoom in.
    }
    \label{fig:vis}
\end{figure*}

\noindent \textbf{Nearest-neighbor query.}
In Fig.~\ref{fig:vis}, we visualize the $8$ predicted nearest neighbors, from GCD~\cite{gcd} and our PromptCAL, of $20$ randomly selected query images, which are labeled with correct (green) and incorrect (red). 
Specifically, we first randomly sample a subset from ImageNet-1K, and conduct KNN search (with cosine distance) for given random queries in {\tt [CLS]} embedding space.
We can observe that PromptCAL generally exhibits higher retrieval precision (\eg, for ``n02006656'' in $3^{rd}$ row, ``02018207'' in $5^{th}$ row, ``n02027492'' in $8^{th}$ row). To summarize, our PromptCAL learns more semantically calibrated local structures. We also notice that both GCD and PromptCAL fails on ``n01695060'' in $11^{th}$ row, which, we guess, is due to the confusing view angle of the query image and high visual similarities between lizards of different species. \par

\section{Efficiency analysis}
\label{app:efficiency}
Compared with the raw ViT backbone (GCD~\cite{gcd}), our PromptCAL only adds negligible computation overheads during inference, since the only overheads origin from visual prompts. In Table~\ref{tab:time}, we quantitatively list inference time per image, thoughput, and FLOPs for PromptCAL. It can be observed that our PromptCAL achieves comparable inference efficiency with the raw ViT backbone. \par

\begin{table}[]
    \centering
    \resizebox{\linewidth}{!}{
    \begin{tabular}{c|ccc}
    \toprule
        Method & Time (s/per img) & Throughput (img/s) & FLOPs \\
    \midrule
        GCD~\cite{gcd} & $1.70 \times 10^{-3}$ & 586 & 35.1 \\
        \rowcolor{Green}
        PromptCAL (our) & $1.79 \times 10^{-3}$ & 558 & 36.1 \\
    \bottomrule
    \end{tabular}
    }
    \caption{\textbf{Comparison on inference time, throughput, and FLOPs based on ViT-B/16 backbone.}}
    \label{tab:time}
\end{table}

\section{Broader impact and limitations}
\label{app:limitations}
It should be noticed that although our method achieves state-of-the-art performance on generalized novel category discovery problem, the performance gap between the fully supervised counterpart and our method still exists.
Besides, in real world, the data can be more complicated and uncurated. For instance, realistic data may follow long-tail distributions, human-annotation may incur noises, and the vocabulary maybe huge. We leave these for future research. \par

\section{License for experimental datasets}
\label{app:lic}
All datasets used in our experiments are permitted for research use. CIFAR-100 and CIFAR-10~\cite{cifar} are released under MIT license for research use. 
ImageNet-100, the subset of ImageNet~\cite{imagenet}, also allows for research purpose.
Besides, CUB-200~\cite{dataset-cub}, Aircraft~\cite{aircraft}, StanfordCars~\cite{scars} also permits for research purpose.
Herbarium19~\cite{herb} are released for non-commercial purposes. \par

\begin{center}
\begin{algorithm*}[p]
\algsetup{linenosize=\tiny}
\DontPrintSemicolon
\small

  \KwInput{Training dataset $\mathcal{D}=\mathcal{D}_u \cup \mathcal{D}_l$, an ImageNet pre-trained ViT backbone $f(\cdot | \theta)$, and a randomly-initialized {\tt [CLS]} projection head $g(\cdot | \theta_{\text{H}})$.
  }
  
  \KwOutput{Trained prompt-adapted model $f(\cdot | \theta, \theta_{\text{P}})$.
  }
  
  \text{Initialize prompt-adapted backbone with random prompts into $f(\cdot | \theta, \theta_{\text{P}})$.} \par
  \text{Randomly initialize prompt projection head $g_{\text{P}}(\cdot | \theta_{\text{P},\text{H}})$ from $g$.} \par 
  
  \tcc{\textbf{\color{cyan}Stage 1: Warm-up Training}}
    \For{\text{each epoch} $e$=1...$E_1$ }{
      \For{\text{each batch} $(\mathbf{X}, \mathbf{Y}) \in \mathcal{D}$}{
          $\mathbf{Z}, \overline{\mathbf{Z}}_{\text{P}} = \text{Forward}(\mathbf{X}, f, g, g_{\text{P}})$ \tcp{forward backbone and heads} \par 
          Compute overall SemiCL loss $L_1$ by Eq.~(2) on $\mathbf{Z}, \overline{\mathbf{Z}}_{\text{P}}$. \par 
          Back-propagation and optimize $\theta, \theta_{\text{P}}, \theta_{\text{H}}, \theta_{\text{P}, \text{H}}$. \par 
      }
  }
  
  \tcc{\textbf{\color{cyan}Stage 2: Contrastive Affinity Learning}}
  \text{Initialize memory $\mathcal{M}, \mathcal{M}_{\text{P}}$}. \par 
  \text{Initialize teacher $f_T, g_T, g_{\text{P}, T}$ from the student model}. \par
  \For{each epoch $e$=1...$E_2$}{
      \For{\text{each batch} $(\mathbf{X}, \mathbf{Y}) \in \mathcal{D}$}{
          \tcc{\color{cyan}Forward} \par 
          $\mathbf{H}, \overline{\mathbf{H}}_{\text{P}}, \mathbf{Z}, \overline{\mathbf{Z}}_{\text{P}} = \text{Forward}(\mathbf{X}, f, g, g_{\text{P}})$ \tcp{forward student} \par 
          $\mathbf{H}_T, \overline{\mathbf{H}}_{\text{P}, T}, \mathbf{Z}_T, \overline{\mathbf{Z}}_{\text{P}, T} = \text{Forward}(\mathbf{X}, f_T, g_T, g_{\text{P}, T})$ \tcp{forward teacher} \par 
          
          \tcc{\color{cyan}SemiAG for {\tt [CLS]}}
          Concatenate embedding $E \leftarrow [\mathbf{H}_T; \mathcal{M}]$ for {\tt [CLS]} token and construct sub-graph $\mathbf{G}_{\mathcal{H}}^\prime$. \par
          Compute binarized affinity graph $\mathbf{G}_b^\prime$ from $\mathbf{G}_{\mathcal{H}}^\prime$ by applying SemiAG in Eq.~(4) (5) (6) sequentially. \par
          Obtain pseudo positives $\mathcal{P}_a$ and pseudo anchors $\mathcal{A}_a$ from $\mathbf{G}_b^\prime$. \par
          Compute CAL loss $L_{\text{CAL}}^{\text{CLS}}$ for {\tt [CLS]} with $\mathcal{P}_a$ and $\mathcal{A}_a$ on $\mathbf{H}$ by Eq.~(7). \par 
          
          \tcc{\color{cyan}SemiAG for {\tt [P]}, similar process to {\tt [CLS]}}
          Concatenate embedding $E_{\text{P}} \leftarrow [\overline{\mathbf{H}}_{\text{P}, T}; \mathcal{M}_{\text{P}}]$ for {\tt [P]} token and construct sub-graph $\mathbf{G}_{\text{P}, \mathcal{H}}^\prime$. \par
          Compute $\mathbf{G}_{\text{P}, b}^\prime$ from $\mathbf{G}_{\text{P}, \mathcal{H}}^\prime$ by applying Eq.~(4) (5) (6) sequentially. \par
          Obtain pseudo labels $\mathcal{P}_{\text{P}, a}$ and $\mathcal{A}_{\text{P}, a}$ from $\mathbf{G}_{\text{P}, b}^\prime$. \par
          Compute CAL loss $L_{\text{CAL}}^{\text{P}}$ for {\tt [P]} with $\mathcal{P}_{\text{P}, a}$ and $\mathcal{A}_{\text{P}, a}$ on $\overline{\mathbf{H}}_{\text{P}, T}$ by Eq.~(7). \par

          \tcc{\color{cyan}SemiCL loss}
          Compute $L_{\text{sup}}^{\text{CLS}}, L_{\text{self}}^{\text{CLS}}$ for {\tt [CLS]} and $L_{\text{sup}}^{\text{P}}, L_{\text{self}}^{\text{P}}$ for {\tt [P]} on $\mathbf{Z}$ and $\mathbf{Z}_T$ by Eq.~(8). \par 
          
          \tcc{\color{cyan}Compute total loss}
          Compute {\tt [CLS]} total loss $L_2^{\text{CLS}}$ with $L_{\text{sup}}^{\text{CLS}}, L_{\text{self}}^{\text{CLS}}, L_{\text{CAL}}^{\text{CLS}}$ by Eq.~(9). \par
          Compute overall total loss $L_2$ with $L_2^{\text{CLS}}$ and its DPR counterpart $L_2^{\text{P}}$ by Eq.~(10). \par
          
          \tcc{\color{cyan}Back propagation}
          Back-propagation and optimize student $\theta, \theta_{\text{P}}, \theta_{\text{H}}, \theta_{\text{P}, \text{H}}$. \par 
          $\mathcal{M} \leftarrow \text{Enqueue}(\mathcal{M}, \mathbf{H}_T)$, $\mathcal{M}_{\text{P}} \leftarrow \text{Enqueue}(\mathcal{M}_{\text{P}}, \overline{\mathbf{H}}_{\text{P}, T})$ \tcp{update memories} \par 
          Update momentum teacher with current student. \par
      }
  }
  \Return{$f(\cdot | \theta, \theta_{\text{P}})$} \par
\caption{\mbox{PromptCAL} training algorithm.}
\label{algo}
\end{algorithm*}
\end{center}


\end{document}